\documentclass[11pt]{article}

\usepackage[final]{acl}
\usepackage{times}
\usepackage{latexsym}
\usepackage[T1]{fontenc}
\usepackage[utf8]{inputenc}
\usepackage{microtype}
\usepackage{inconsolata}
\usepackage{graphicx}

\usepackage{amsmath}
\usepackage{amssymb}
\usepackage{booktabs}
\usepackage{colortbl}
\usepackage{float}
\usepackage{caption}
\usepackage{subcaption}
\usepackage{listings}

\definecolor{highlightrow}{HTML}{FFF4D6}

\usepackage{paralist}
\usepackage[most]{tcolorbox}
\usepackage{hyperref}
\definecolor{linkblue}{HTML}{000099}
\hypersetup{colorlinks=true,linkcolor=linkblue,citecolor=linkblue,urlcolor=linkblue}
\usepackage{cleveref}

\newtcolorbox{promptbox}[1]{
  enhanced, breakable,
  colback=gray!4, colframe=gray!55!black, boxrule=0.5pt,
  title={#1}, fonttitle=\bfseries\small, coltitle=white,
  attach boxed title to top left={yshift=-2mm,xshift=4mm},
  boxed title style={colback=gray!55!black,boxrule=0pt,sharp corners},
  arc=2pt,
  before skip=8pt, after skip=8pt,
  left=8pt, right=8pt, top=10pt, bottom=8pt,
}

\title{Fluid Reasoning Representations}
\author{%
  \textbf{Dmitrii~Kharlapenko$^{1,2}$\textsuperscript{*}} \quad \textbf{Terry~Jingchen~Zhang$^{1,2}$\textsuperscript{*}} \quad \textbf{Arth~Singh$^{1,3}$\textsuperscript{*}} \\
  \textbf{Alessandro~Stolfo$^{2}$} \quad \textbf{Arthur~Conmy} \quad \textbf{Mrinmaya~Sachan$^{2}$} \quad \textbf{Zhijing~Jin$^{1,4}$} \\
  $^{1}$Jinesis Lab, University of Toronto\&Vector Institute \\
  $^{2}$ETH Zurich \quad $^{3}$AIM Intelligence \\
  $^{4}$Max Planck Institute for Intelligent Systems, T\"ubingen, Germany \\
  \textsuperscript{*}Equal contribution \\
}

\begin{document}
\maketitle
    
\begin{abstract}
Frontier large language models increasingly solve complex tasks involving abstract concepts through extended test-time thinking.
Yet we lack a mechanistic account of how extended thinking changes hidden-state representations over the course of a reasoning trace.
We introduce \textit{Fluid Reasoning Representations} (FRRs), a representation-level account of how LLMs organize action and predicate concepts during self-generated reasoning, and test them on obfuscated planning, symbolic, and mathematical tasks where task-relevant words are replaced while problem structure is preserved.
Across open-weight base, instruction-tuned, and extended-thinking LLMs, representations of the same action or predicate become more similar across wordings and move toward the corresponding unobfuscated concepts over the reasoning trace.
Causal probes show that these representations affect behavior: cross-naming steering improves held-out accuracy beyond Gaussian and shuffled controls, symbolic patching retains more action information than shuffled patching, and subtracting refined directions degrades accuracy; together, these results suggest that extended thinking strengthens a representation dynamic already present at lower magnitude in base and instruction-tuned LLMs.
Our codebase is open-sourced \href{https://github.com/AI4Collaboration/Fluid-Reasoning-Representation}{here}.
\end{abstract}

\section{Introduction}

Test-time thinking has become a central feature for frontier large language models (LLMs) to tackle complex tasks involving abstract concepts. Scaling inference-time computation can improve LLM performance, in some settings more effectively than increasing model size~\citep{snell2024scalingllmtesttimecompute}. Yet we still lack a mechanistic account of what this process changes inside the model~\citep{venhoff2025understandingreasoningthinkinglanguage,xu2025largereasoningmodelssurvey}. Prior work shows that task and function information can be encoded, localized, and manipulated in activation space~\citep{todd2024function,hendel-etal-2023-context,park2025iclrincontextlearningrepresentations}. This representation-level view raises a sharper question: does extended thinking mainly allocate additional computation to search and template reuse, or does it reshape internal representations used by later reasoning steps?

We address this problem by studying model representations on obfuscated reasoning tasks. These tasks keep the underlying problem structure fixed while systematically replacing task-relevant words, making it possible to test whether representations of the same operation become more consistent across wordings and useful for downstream behavior. We study this setting across planning, symbolic, and mathematical reasoning tasks and multiple open-weight LLMs, including QwQ, Qwen2.5, DeepSeek-R1-Distill-Qwen, Nemotron, and Seed-OSS variants; controlled planning domains serve as the primary causal testbed. The detailed task construction, representation extraction, and intervention setup are given in \Cref{sec:methodology}; the introduction focuses on the central question of whether test-time thinking changes the representations that support reasoning.

We introduce \textbf{Fluid Reasoning Representations} (FRRs), a representation-level account of how LLMs organize action and predicate concepts during self-generated reasoning. FRRs describe a dynamic in which representations of the same action or predicate become more similar across wordings and move toward the corresponding unobfuscated concepts. We test this account by asking three questions: whether equivalent concepts become more consistent across different wordings, whether they partially align with the corresponding unobfuscated concepts, and whether interventions on these representations change downstream reasoning behavior.

\noindent\textbf{Contributions.} This paper contributes a mechanistic account of thinking behavior in LLMs by defining FRRs and showing how they can be measured in obfuscated reasoning traces. Empirically, action and predicate representations become more similar across wordings over the reasoning trace and move toward corresponding unobfuscated concepts. Causal probes show that these representations affect behavior: cross-naming steering improves held-out accuracy beyond Gaussian and shuffled controls, symbolic patching retains more action information than shuffled patching, and subtracting refined directions degrades accuracy. Base and instruction-tuned LLMs exhibit the same dynamic at lower magnitude, suggesting that extended thinking strengthens an existing representational dynamic rather than creating it from scratch; alignment with unobfuscated concepts still does not guarantee functional interchangeability.

\begin{figure*}[!ht]
    \centering
    \includegraphics[width=0.8\textwidth]{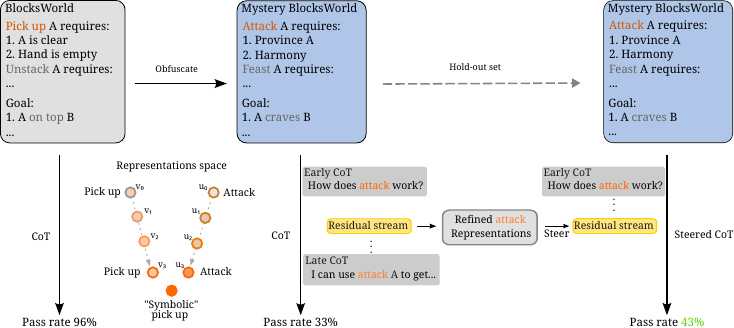}
    \caption{\textbf{Pipeline overview.} We extract action/predicate representations from obfuscated reasoning traces and test them with steering interventions.}
    \label{fig:main_diagram}
\end{figure*}

\section{Related Work}

\noindent\textbf{Representation-level accounts of LLM behavior.} Prior work shows that LLM behavior can often be understood through activation-space directions or features\footnote{\url{https://transformer-circuits.pub/2021/framework/index.html}} \citep{nanda-etal-2023-emergent,olah2024lrh}, and that task/function information can be localized, manipulated, or constructed in context \citep{todd2024function,hendel-etal-2023-context,park2025iclrincontextlearningrepresentations}. These studies make representation analysis useful for causal questions, but usually focus on fixed tasks, prompt-level task vectors, or short-context adaptation. We instead ask how self-generated test-time thinking changes action and predicate representations over the course of a trace, and whether those changing representations matter for later behavior.

\noindent\textbf{Thinking behavior and internal mechanisms.} Recent reasoning interpretability work studies behavioral and representational structure in extended thinking traces, including uncertainty, backtracking, verification, and reasoning-specific internal features \citep{venhoff2025understandingreasoningthinkinglanguage,bogdan2025thoughtanchorsllmreasoning,lee2025geometryselfverificationtaskspecificreasoning,ward2025reasoningfinetuningrepurposeslatentrepresentations}. This work helps explain what appears in traces and where some internal features are located. FRRs target a complementary question: whether the thinking process reshapes task representations that the model later uses, which we test by comparing representations across timestamps and intervening on them.

\noindent\textbf{Obfuscated reasoning.} Obfuscated reasoning tasks keep the underlying problem structure fixed while replacing task-relevant words, exposing how much LLMs rely on familiar lexical cues \citep{valmeekam2023planbenchextensiblebenchmarkevaluating,valmeekam2024llmscantplanlrms}. We use this setting as a controlled identification strategy: vary surface namings while preserving the same task, then test whether representations for corresponding actions and predicates become more consistent, align with clean-domain concepts, and causally affect behavior. This differs from robustness evaluation because the goal is not only to measure failure under obfuscation, but to isolate the internal adaptation that makes some obfuscated reasoning succeed.

\section{Methodology}
\label{sec:methodology}

\noindent\textbf{Dataset.} We use BlocksWorld, a classic planning domain \citep{ipc1998}, as the base environment: each puzzle specifies initial and goal block arrangements, and the agent must use \textit{pick-up}, \textit{put-down}, \textit{stack}, and \textit{unstack} under simple physical constraints. We generate and verify puzzles with PlanBench \citep{valmeekam2023planbenchextensiblebenchmarkevaluating} (prompt in \Cref{app:prompt_blocksworld}). To create the obfuscated setting, Mystery BlocksWorld replaces every action and predicate with a semantically unrelated word under a complete \textbf{naming}, preserving logical structure while breaking familiar surface semantics (prompt in \Cref{app:prompt}). We generated 14 variants beyond the original, yielding 15 obfuscations of the same domain; a \textbf{puzzle} is a unique initial-goal state pair, and our analysis uses 300 four-block puzzles mapped across all 15 namings.

\noindent\textbf{Baseline evaluations.} QwQ-32B is the strongest open-weight LLM in our evaluation, solving regular BlocksWorld at 96\% and Mystery BlocksWorld at 35\%; successful mystery solutions typically require 15--20k-token traces (\Cref{tab:blocksworld_performance}). A reasoning-behavior breakdown following \citet{venhoff2025understandingreasoningthinkinglanguage} appears in \Cref{app:behavior}.

\begin{table*}[t]\small
\centering
\caption{BlocksWorld and \textbf{Naming 1} Mystery BlocksWorld performance on 300 puzzles. ``Tokens'' is average thinking-trace length; ``Accuracy Preserved'' is Mystery/Standard. $^*$ indicates unparsable formatting.}
\begin{tabular}{lcccccc}
\toprule
\textbf{LLM} & \multicolumn{2}{c}{\textbf{BlocksWorld}} & \multicolumn{2}{c}{\textbf{Mystery}} & \textbf{Accuracy Preserved} \\
\cmidrule(lr){2-3} \cmidrule(lr){4-5}
& \texttt{Acc} & \texttt{Tokens} & \texttt{Acc} & \texttt{Tokens} & \\
\midrule
\multicolumn{6}{l}{\textbf{Regular LLMs}} \\
\midrule
GPT-4.1 (step-by-step) & 0.92 & 556 & 0.18 & 3837 & \textbf{20\%} \\
Qwen2.5-32B-base & 0.21 & 71 & 0.00 & 1390 & \textbf{0\%} \\
Qwen2.5-32B-instruct (step-by-step) & 0.38 & 353 & 0.00 & 1479 & \textbf{0\%} \\
Llama-3.3-70B-instruct (step-by-step) & 0.40 & 760 & 0.02 & 1142 & \textbf{5\%} \\
\midrule
\multicolumn{6}{l}{\textbf{LLMs with Extended Test-Time Thinking}} \\
\midrule
DeepSeek-R1-Distill-Qwen-32B & 0.81 & 2387 & 0.08 & 8500 & \textbf{10\%} \\
DeepSeek-R1-Distill-Llama-70B & 0.66 & 2674 & 0.10 & 10636 & \textbf{15\%} \\
Nemotron-49B-instruct & 0.48$^*$ & 1162 & 0.19 & 9200 & \textbf{40\%} \\
\rowcolor{highlightrow}\textbf{QwQ-32B} & \textbf{0.96} & 3633 & \textbf{0.35} & 16186 & \textbf{36\%} \\
\bottomrule
\end{tabular}
\label{tab:blocksworld_performance}
\end{table*}

\noindent\textbf{Naming variation.} QwQ-32B accuracy varies across namings (0.05--0.47): coherent alternative domains and reversible-sounding operations are hardest, while abstract or mismatched replacements are easier (full results in \Cref{app:mystery_performance}). \textit{Mystery naming 3} uses random strings; QwQ-32B often recognizes BlocksWorld and maps symbols quickly in this condition (about 2k tokens vs. 15--20k elsewhere), consistent with training-set exposure, so we exclude naming 3 from representational analyses because it largely bypasses the semantic adaptation process.

\subsection{Representation Collection}
\label{sec:representation_collection}

\noindent\textbf{Representation construction.} Following \citet{park2025iclrincontextlearningrepresentations}, we extract action and predicate vectors from LLM activations during BlocksWorld and Mystery BlocksWorld reasoning. For traces $\mathcal{B}$, phrase $a$, naming $N$, layer $L$, and timestamp $T$, we scan the 100-token window before $T$, collect token spans encoding $a$ plus one preceding token, average hidden states within each span and then across spans, and repeat every 200 tokens at all layers:
\begin{equation}
    \mathbf{r}_a^{N,L,T} = \frac{1}{|\mathcal{S}|} \sum_{s \in \mathcal{S}} \left( \frac{1}{|s|} \sum_{i \in s} \mathbf{h}_i^L \right),
\end{equation}
where $\mathcal{S}$ is the set of matched spans. Within each naming, we follow \citet{venhoff2025understandingreasoningthinkinglanguage} and center over the action or predicate set $\mathcal{A}$:
\begin{equation}
\label{eq:centered}
    \tilde{\mathbf{r}}_a^{N,L,T} = \mathbf{r}_a^{N,L,T} - \frac{1}{|\mathcal{A}|} \sum_{a' \in \mathcal{A}} \mathbf{r}_{a'}^{N,L,T},
\end{equation}
which removes the per-naming common-mode component. We then average centered vectors across namings to isolate what is shared across surface forms:
\begin{equation}
\label{eq:cross_naming}
    \bar{\mathbf{r}}_a^{L,T} = \frac{1}{|\mathcal{N}|} \sum_{N \in \mathcal{N}} \tilde{\mathbf{r}}_a^{N,L,T},
\end{equation}
where $\mathcal{N}$ is the set of namings. We validate this construction with PCA, clean-domain similarity, positive steering, and negative steering in \Cref{sec:rep_avg,sec:steering}.

\subsection{Fluid Reasoning Representations}
\label{def:frr}
An LLM exhibits \emph{Fluid Reasoning Representations} for action set $\mathcal{A}$ if, for some layer $L$, $\{\mathbf{r}_a^{N,L,T}\}_{a,N,T}$ satisfies three criteria as $T$ increases:
\begin{compactenum}
    \item[\textbf{C1}] \emph{Cross-naming alignment.} The cosine similarity between $\tilde{\mathbf{r}}_a^{N,L,T}$ and $\bar{\mathbf{r}}_a^{L,T}$ increases with $T$, while the similarity between $\tilde{\mathbf{r}}_a^{N,L,T}$ and $\bar{\mathbf{r}}_{a'}^{L,T}$ for $a'\!\neq\!a$ does \emph{not} (and ideally decreases). [Tested: \Cref{fig:representational_convergence}.]
    \item[\textbf{C2}] \emph{Clean-domain similarity.} The cross-naming vector $\bar{\mathbf{r}}_a^{L,T}$ approaches the centered representation of the same action or predicate computed on the unobfuscated domain, despite no token overlap. [Tested: \Cref{fig:representational_convergence}.]
    \item[\textbf{C3}] \emph{Causal contribution.} Injecting $\bar{\mathbf{r}}_a^{L,T}$ into early reasoning improves held-out accuracy more than (a) Gaussian noise of matched norm, (b) shuffled assignments of actions to vectors, and (c) in-naming vectors $\tilde{\mathbf{r}}_a^{N,L,T}$; subtracting the refined direction degrades accuracy beyond a shuffled control of equal magnitude. [Tested: \Cref{sec:steering_positive,sec:steering_negative}; \Cref{tab:steering_improvements}.]
\end{compactenum}
\noindent\textbf{Relation to prior work.} Unlike explicit in-context remappings in synthetic graph tasks \citep{park2025iclrincontextlearningrepresentations}, our setting requires inference from $\sim$15--20k self-generated thinking tokens and adds clean-domain alignment (C2) plus causal tests (C3).

\section{Representational Studies}
\label{sec:rep}
In this section, we first test whether QwQ-32B \citep{qwq32b} refines action and predicate representations during extended Mystery BlocksWorld reasoning. We operationalize \textit{Fluid Reasoning Representations} as increasing alignment between in-naming representations and their counterparts averaged across namings, plus causal effects under intervention.

\subsection{Representational Convergence}
\label{sec:rep_convergence}
\begin{figure}[t]
    \centering
    \includegraphics[width=\linewidth]{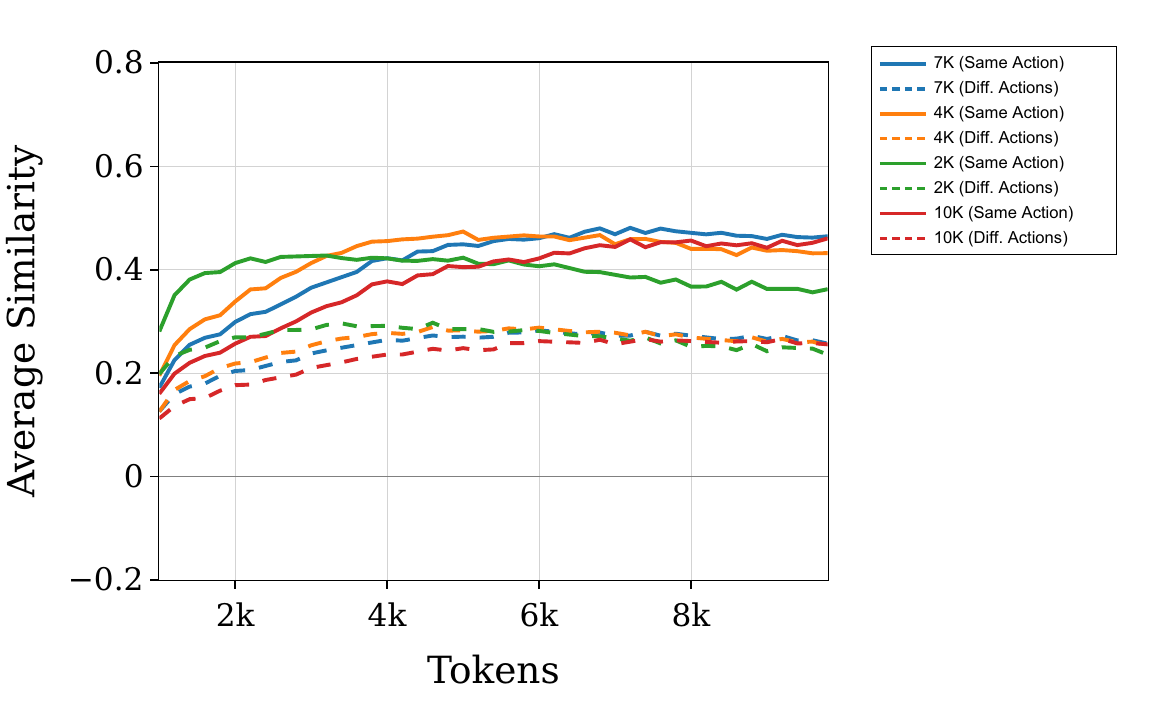}
    \caption{\textbf{Action and predicate representations converge across obfuscated namings over the reasoning trace.} Cross-naming cosine similarity grows over tokens and plateaus near 7k; same-action pairs separate from different-action pairs.}
    \label{fig:diff_timestamps}
\end{figure}

\begin{figure}[t]
    \centering
    \includegraphics[width=\linewidth,trim={0 0 45pt 0},clip]{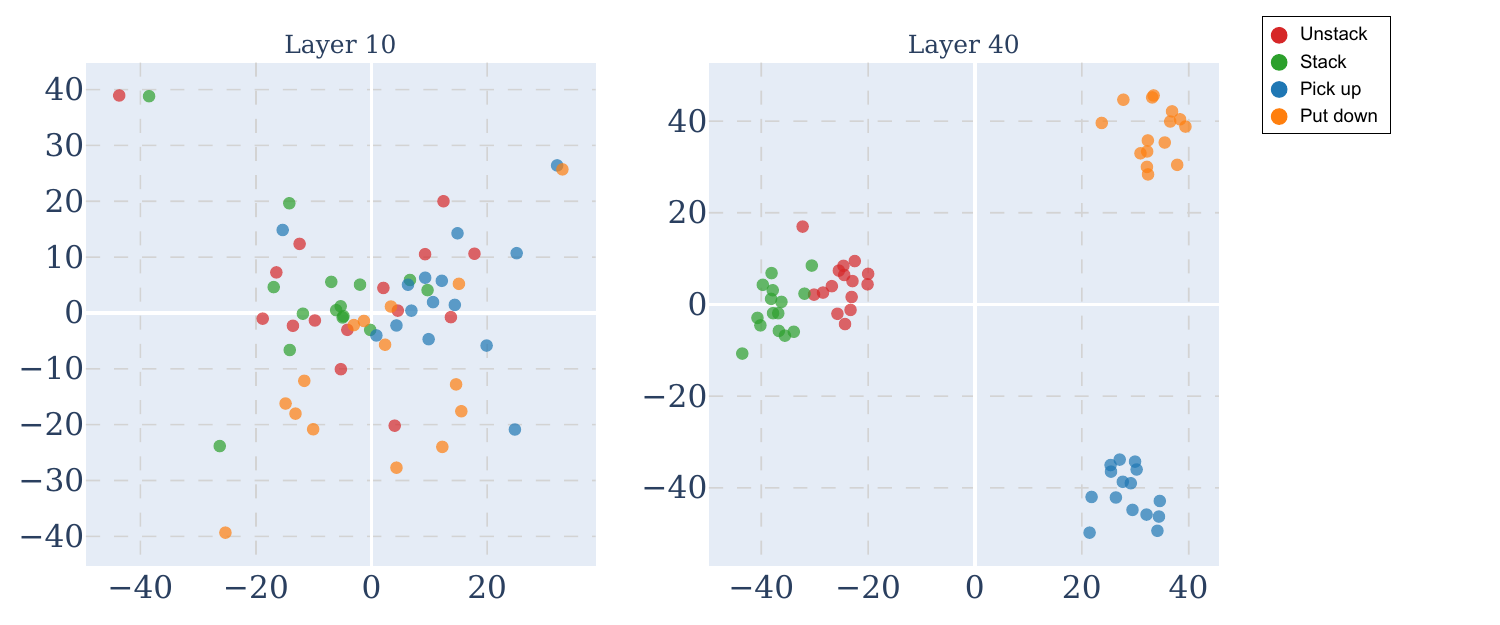}
    \caption{\textbf{Layer-wise PCA at 7k tokens shows clean action-type clustering.} Equivalent actions cluster across obfuscated namings, and the separation sharpens with depth.}
    \label{fig:pca_7k}
\end{figure}

We extract in-naming representations from Mystery naming 1 at 2k, 4k, 7k, and 10k tokens and compare them with representations from all other namings. \Cref{fig:diff_timestamps} shows that corresponding actions become more similar across namings during reasoning, plateauing around 7k tokens, while different-action similarities remain lower; \Cref{fig:pca_7k} shows the same pattern in PCA space, with equivalent actions clustering across namings especially in deeper layers. The residual similarity among different actions is mainly driven by ``stack'' and ``unstack'' being closer to each other than to ``pick up'' or ``put down''.
This convergence also clarifies the cost of reasoning. QwQ-32B solves Standard BlocksWorld with $\sim$3.6k tokens, but successful Mystery solutions require 15--20k tokens (Table~\ref{tab:blocksworld_performance}). The first $\sim$7k tokens align with the convergence phase in \Cref{fig:diff_timestamps}, suggesting that representational alignment, not raw token count alone, is the relevant cost.

\subsection{Representation Similarity}
\label{sec:rep_avg}
\begin{figure*}[h!]
    \centering
    \begin{subfigure}[b]{0.48\textwidth}
        \centering
        \includegraphics[width=\linewidth]{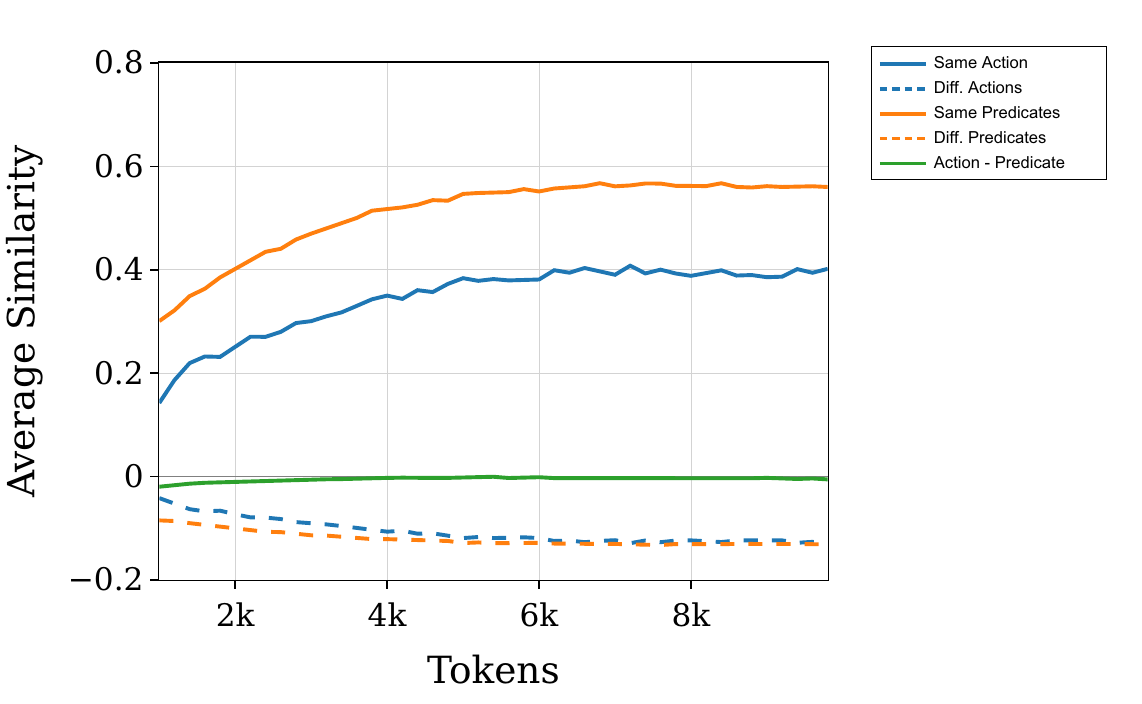}
        \label{fig:sims_avg}
    \end{subfigure}
    \hfill
    \begin{subfigure}[b]{0.48\textwidth}
        \centering
        \includegraphics[width=\linewidth]{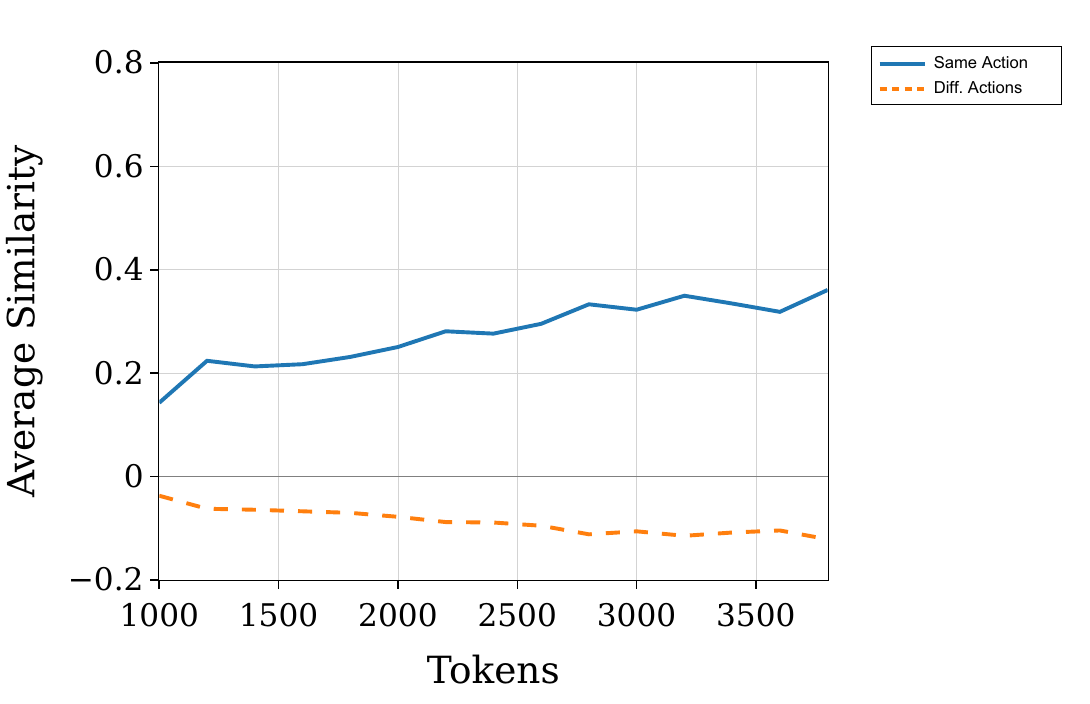}
        \label{fig:sims_clean}
    \end{subfigure}
    \caption{\textbf{Both Mystery and clean BlocksWorld traces drift toward the same cross-naming averages.} (Left) Centered Mystery representations approach their cross-naming average over the trace, while different-action pairs separate. (Right) Clean BlocksWorld representations also move toward the cross-naming average, linking obfuscated adaptation to the corresponding unobfuscated task concepts; predicates are omitted because their tokens are harder to identify in clean traces.}
    \label{fig:representational_convergence}
\end{figure*}

We next compare naming-specific representations with cross-naming averages and clean-domain representations.
Similarity to the corresponding cross-naming average increases during reasoning and plateaus around 7k tokens, while different-action similarities become increasingly negative (\Cref{fig:representational_convergence}). Clean BlocksWorld representations also move toward the same cross-naming vectors over time, linking the obfuscated-trace dynamic to the model's representation of the unobfuscated task.
Together, these comparisons give the cross-naming average a concrete interpretation: it tracks a role-specific direction that preserves action identity while abstracting over surface labels. The clean-domain comparison anchors the same direction to the unobfuscated action concept, linking the obfuscated representations to the task structure that the model must use to plan. We therefore treat the cross-naming average as an empirical construction of shared role information.

\subsection{Qwen2.5 Trace-Processing Controls}
\begin{figure*}[t]
    \centering
    \begin{subfigure}[b]{0.48\textwidth}
        \centering
        \includegraphics[width=\linewidth]{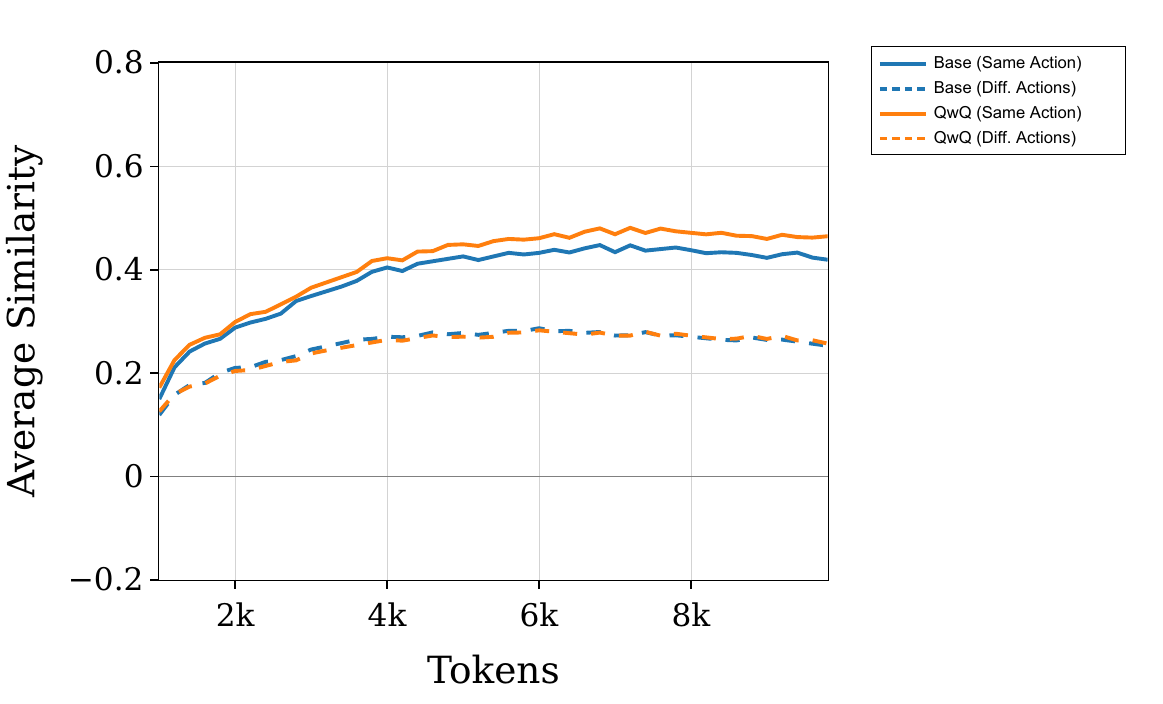}
        \label{fig:sims_pair_base_qwq}
    \end{subfigure}
    \hfill
    \begin{subfigure}[b]{0.48\textwidth}
        \centering
        \includegraphics[width=\linewidth]{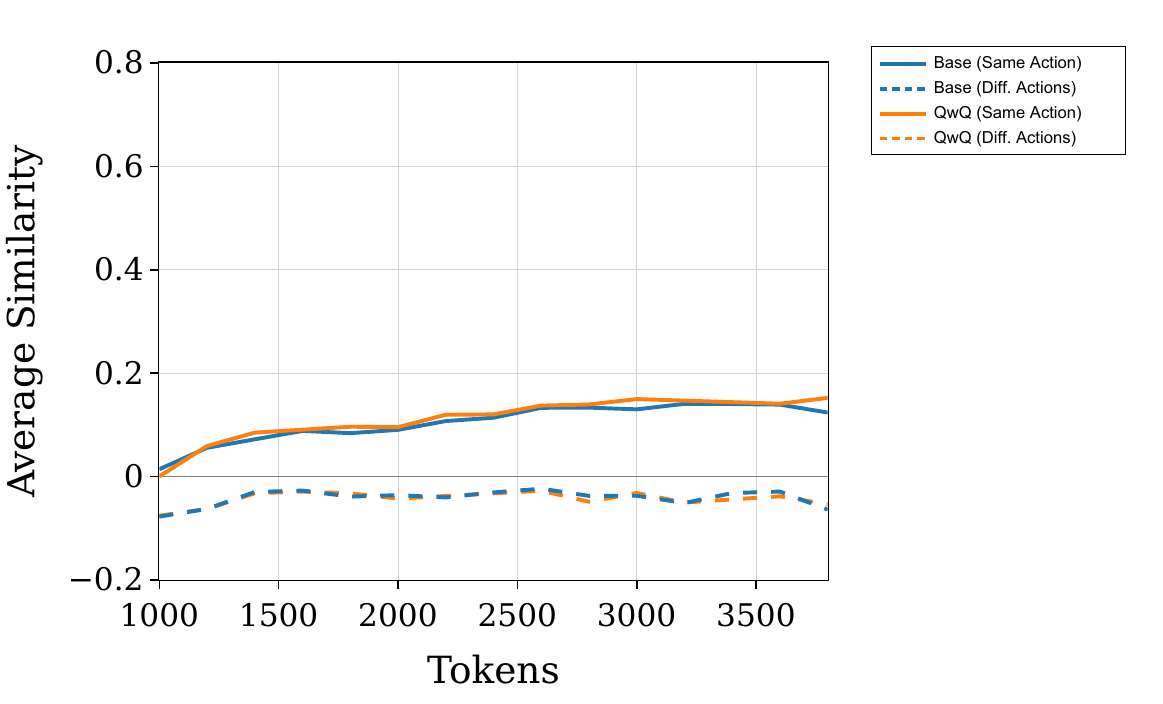}
        \label{fig:sims_pair_base_qwq_from_clean}
    \end{subfigure}
    \caption{\textbf{Qwen2.5 base and instruct exhibit weaker but qualitatively similar FRR dynamics when fed QwQ-generated traces} (7k tokens). (Left) Cross-naming similarity vs.~QwQ. (Right) Clean-domain similarity vs.~QwQ. Both controls reproduce the trend at lower magnitude, indicating extended-thinking post-training amplifies the dynamic rather than creates it.}
    \label{fig:base_model_comparison}
\end{figure*}

We tested whether the FRR phenomenon is specific to post-training for extended test-time thinking by feeding the \emph{same} QwQ-generated traces through Qwen2.5-32B-instruct and Qwen2.5-32B-base,\footnote{\url{https://huggingface.co/Qwen/Qwen2.5-32B-Instruct}} then extracting representations identically. Both Qwen2.5 variants exhibit qualitatively similar but weaker convergence dynamics (\Cref{fig:base_model_comparison}), with the base LLM adapting more slowly, consistent with the fact that it is being asked to process traces it would not itself generate. This comparison places FRRs on a broader in-context representational adaptation substrate \citep{park2025iclrincontextlearningrepresentations} that is already present in base and instruction-tuned Qwen2.5 LLMs. The contrast also matters: models trained to sustain extended test-time thinking generate the self-consistent traces that drive alignment past the 7k-token plateau (Table~\ref{tab:blocksworld_performance}) and make the dynamic more visible under intervention. We therefore frame FRRs as a representational dynamic that is \textbf{amplified and operationally exploited} by extended test-time thinking rather than created from scratch.

\subsection{Cross-Model/Cross-Domain Analysis}
We also study the convergence dynamic on three additional open-weight LLMs: DeepSeek-R1-Distill-Qwen-32B \citep{deepseekai2025deepseekr1incentivizingreasoningcapability}, Nemotron-49B-instruct,\footnote{\url{https://huggingface.co/nvidia/Llama-3_3-Nemotron-Super-49B-v1}} and Seed-OSS-36B-instruct \citep{seed2025seed-oss}. All three reproduce both the increasing in-naming/cross-naming alignment of criterion~\textbf{C1} (\Cref{app:cross_model_similarities}) and the layer-wise PCA separation of action types (\Cref{fig:pca_7k_all}). Across our 20 namings, which span coherent alternative domains (gardening, legal, cooking), mystical framings, and nonsense strings (\Cref{app:variants}), baseline accuracy ranges 0.05--0.65 with naming, but C1 holds for every non-degenerate naming and the cross-naming construction in \Cref{eq:cross_naming} generalizes to namings withheld from steering-vector construction (\Cref{tab:mystery_performance_full}). These results keep the empirical scope focused on structured planning while extending beyond a single naming or LLM.

Finally, to test whether the same representational dynamic holds beyond BlocksWorld, we ran the representation pipeline of \Cref{sec:representation_collection} on a second LLM (Nemotron-49B-instruct) and two additional families: \emph{Mystery Logistics} (PDDL Logistics with action and predicate vocabulary obfuscated under five surface variants) and \emph{GSM8K-Renamed} (math word problems with nouns and operation verbs replaced under four variants, numerals and person names preserved). We collect 200 traces per cell at each LLM's default steering layer (47 for QwQ-32B, 60 for Nemotron-49B-instruct) and report C1 as the mean cross-naming cosine over actions and predicates within each family (\Cref{tab:phase1_c1}). Three conclusions follow from \Cref{tab:phase1_c1}: C1 remains positive and usually strong in every (LLM, family) cell despite zero token overlap across namings, so FRRs transfer from BlocksWorld to PDDL Logistics and GSM8K-style math reasoning rather than reflecting a BlocksWorld-only artifact; the LLM ranking is domain dependent, with QwQ-32B giving the clearest BlocksWorld alignment while Seed-OSS-36B-instruct and DeepSeek-R1-Distill-Qwen-32B lead on Logistics and GSM8K; and the Qwen2.5-32B trajectory shows that post-training reshapes a pre-existing substrate, since Qwen2.5-32B-base and Qwen2.5-32B-instruct already align broadly across domains whereas QwQ-32B specializes the dynamic for BlocksWorld and loses cross-domain alignment. Cross-domain mean-residual and C2 analyses support the same picture: the LLMs share a generic structural subspace across families, but cosine alignment alone does not guarantee later functional interchangeability.

These cross-domain results sharpen the scope of the claim. Logistics preserves planning-style state transitions while changing the object and predicate inventory; GSM8K-Renamed preserves arithmetic structure while replacing familiar lexical cues. Positive C1 in both settings shows that the measured alignment extends beyond the specific BlocksWorld vocabulary and actions. The domain-dependent ranking shows that the strength of this alignment depends on both the task family and the LLM, so we use C1 as a diagnostic for consistent role representations within a task family.

\begin{table}[t]
\centering
\footnotesize
\setlength{\tabcolsep}{3pt}
\resizebox{\columnwidth}{!}{%
\begin{tabular}{llrrr}
\toprule
LLM & Family & $|\mathcal{N}|$ & C1$_a$ & C1$_p$ \\
\midrule
\rowcolor{highlightrow}QwQ-32B                  & BlocksWorld & 14 & \textbf{0.94} & \textbf{0.93} \\
Seed-OSS-36B-instruct             & BlocksWorld &  3 & 0.83          & 0.82 \\
Qwen2.5-32B-instruct     & BlocksWorld &  3 & 0.83          & 0.82 \\
DeepSeek-R1-Distill-Qwen-32B  & BlocksWorld &  3 & 0.80          & 0.79 \\
Qwen2.5-32B-base         & BlocksWorld &  3 & 0.79          & 0.75 \\
Nemotron-49B-instruct    & BlocksWorld &  3 & 0.67          & 0.73 \\
\midrule
Qwen2.5-32B-base         & Logistics   &  3 & \textbf{0.84} & 0.81 \\
Qwen2.5-32B-instruct     & Logistics   &  3 & 0.81          & \textbf{0.83} \\
Seed-OSS-36B-instruct             & Logistics   &  3 & 0.77          & 0.80 \\
DeepSeek-R1-Distill-Qwen-32B  & Logistics   &  3 & 0.74          & 0.77 \\
QwQ-32B                  & Logistics   &  3 & 0.68          & 0.55 \\
Nemotron-49B-instruct    & Logistics   &  3 & 0.61          & 0.63 \\
\midrule
Seed-OSS-36B-instruct             & GSM8K       &  2 & \textbf{0.86} & \textbf{0.83} \\
Qwen2.5-32B-base         & GSM8K       &  2 & 0.84          & 0.81 \\
DeepSeek-R1-Distill-Qwen-32B  & GSM8K       &  2 & 0.82          & 0.81 \\
Qwen2.5-32B-instruct     & GSM8K       &  2 & 0.82          & 0.79 \\
Nemotron-49B-instruct    & GSM8K       &  2 & 0.66          & -- \\
QwQ-32B                  & GSM8K       &  2 & 0.65          & -- \\
\bottomrule
\end{tabular}%
}
\caption{\textbf{C1 across (LLM, family) cells.} $|\mathcal{N}|$ is the number of obfuscation namings; C1$_a$ and C1$_p$ are cross-naming cosine alignment for actions and predicates. ``--'' indicates unmeasured predicate alignment; best C1$_a$ per family is bolded.}
\label{tab:phase1_c1}
\end{table}

\section{Causal Validation}
\label{sec:steering}

The representational results suggest that refined vectors encode actionable puzzle structure. We test this with positive steering, symbolic patching, negative steering, and clean-centroid replacement. Together, these probes form a \textbf{causal evidence chain}. Positive steering asks whether a refined vector can help an early trace; symbolic patching tests whether the vector must match the underlying action or predicate; negative steering tests whether the original trace relies on the refined direction; and replacement steering tests how far clean-domain and obfuscated representations can substitute for one another.

\subsection{Positive Steering}
\label{sec:steering_positive}

We inject late-stage action and predicate representations into early reasoning traces and measure held-out accuracy. Gains from cross-naming vectors test whether the representation abstracts beyond a specific lexical naming.

\paragraph{Experimental Setup.} Our steering procedure selects a steering layer $L$, token window $[t_{\text{start}}, t_{\text{end}})$, and steering scale $s$. We collect three types of steering vectors at layer $L$ from the 40 correctly solved puzzles: \textbf{(1)} centered \textbf{in-naming} representations $\tilde{\mathbf{r}}_a^{N,L,T}$ for all actions and predicates, \textbf{(2)} \textbf{cross-naming} representations $\bar{\mathbf{r}}_a^{L,T}$ averaged across all namings, and \textbf{(3)} random Gaussian vectors $\mathbf{v}_{\text{rand}}[a]$ scaled to match the norm of \textbf{in-naming} representations. We extract prefixes of $t_{\text{end}}$ tokens from a hold-out set of 100 different 4-block problem rollouts as our intervention dataset. For each prefix, we identify token indices $i$ corresponding to action or predicate $a$, obtain hidden states $\mathbf{h}_i^L$ at layer $L$, and apply the following norm-preserving intervention:
\begin{align}
    \mathbf{h'}_i^L &= s \cdot \mathbf{h}_i^L + (1-s) \cdot \mathbf{v}_{\text{type}}[a], \\
    \mathbf{h}_i^L &\leftarrow \mathbf{h'}_i^L \cdot \frac{\|\mathbf{h}_i^L\|_2}{\|\mathbf{h'}_i^L\|_2},
\end{align}
where $\mathbf{v}_{\text{type}}[a] \in \{\tilde{\mathbf{r}}_a^{N,L,T}, \bar{\mathbf{r}}_a^{L,T}, \mathbf{v}_{\text{rand}}[a]\}$ depending on the experiment condition. This procedure adds the refined representation while preserving activation magnitude. We measure accuracy improvement on steered puzzles compared to non-steered baseline. We selected scale $s=\frac{2}{3}$ after a sweep on layer 20 using \textbf{in-naming} representations (\Cref{app:hyperparams}; see also \Cref{fig:patching_scales}). The steering window is $[1500, 2500]$.

\begin{figure}[t]
    \centering
    \includegraphics[width=\linewidth]{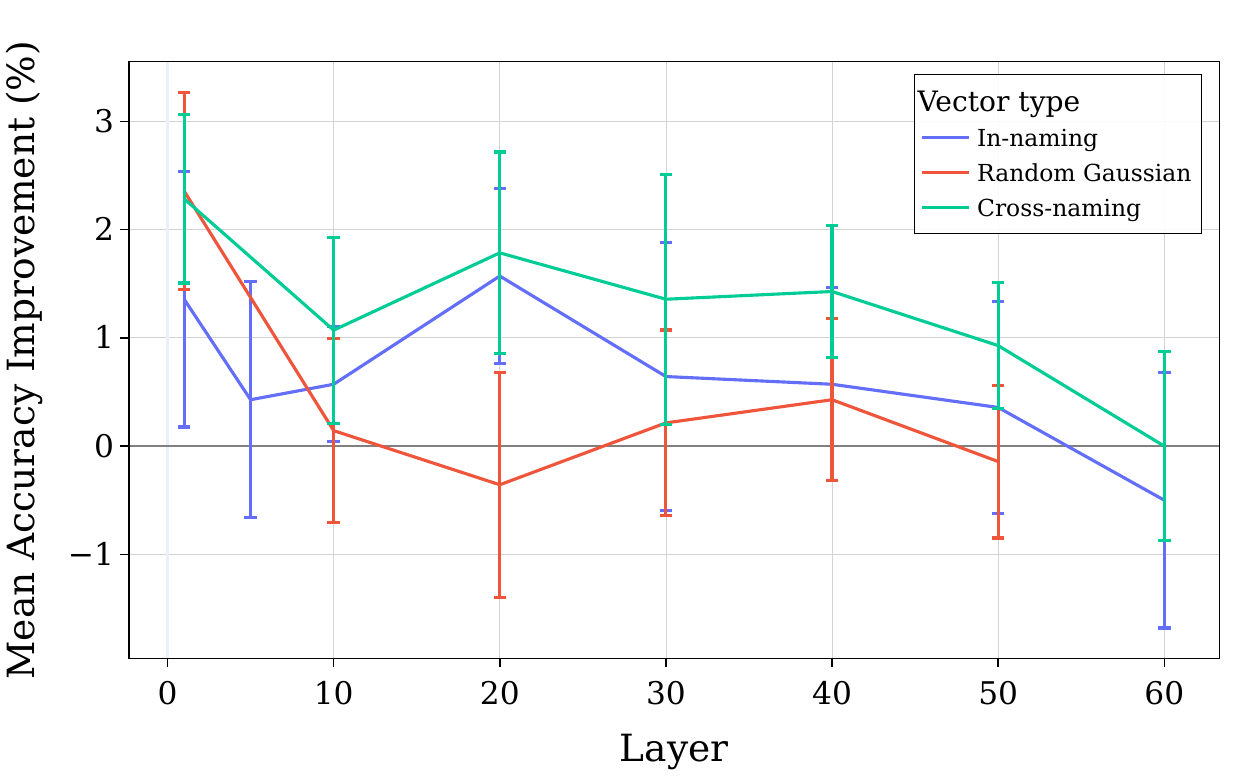}
    \caption{\textbf{Positive steering across layers.} Accuracy change under in-naming, cross-naming, and Gaussian steering; error bars show standard error.}
    \label{fig:positive_steering_layers}
\end{figure}

\paragraph{Results.} \Cref{fig:positive_steering_layers} shows accuracy change after positive steering, averaged across namings (excluding naming 3), as a function of intervention layer $L$, with full per-condition statistics in \Cref{tab:steering_improvements}. The layer-resolved view directly addresses a concern that our gains might come from disrupting misleading surface lexical priors rather than from injecting useful structure: random Gaussian gains are confined to early layers ($L\!\le\!10$); from layer~20 onward, where layer-wise PCA shows the onset of action-type separation across namings (\Cref{app:pca}), random-noise steering stops helping (layer~20 random Gaussian: $-0.36\%$, $p{=}0.627$, \Cref{tab:steering_improvements}), while structured injections continue to yield significant gains (layer~40 cross-naming: $+1.43\%$, $p{=}0.021$). If the cross-naming advantage stemmed from generic surface-token disruption, both random noise and shuffled in-naming permutations would match it at deep layers; they do not (\Cref{app:shuffled_steering}). Mean improvements at significant cells are 1.4--1.8\%; per-naming maxima reach +10\% on coherent-domain namings where surface priors are strongest (\Cref{app:mystery_performance}), but heterogeneity across namings and a layer-by-naming interaction keep the cross-naming-averaged effect bounded (per-naming optima in \Cref{tab:mystery_performance_full}). The experiment is designed as a causal probe, so the relevant signal is the \textbf{structured/random asymmetry}: \textbf{cross-naming $>$ in-naming $\gg$ random}, consistent with criterion~\textbf{C3} of \nameref{def:frr}.

\subsection{Symbolic Patching}
\label{sec:steering_patching}

To test whether adapted representations support abstraction across surface forms, we conduct a patching experiment that replaces naming-specific representations with cross-naming ``symbolic'' representations and tests whether the LLM can operate effectively without access to the original naming-specific encodings.

\paragraph{Symbolic Representation Construction.} We construct symbolic representations to be minimally out-of-distribution while capturing abstract structural information. We collect centered \textbf{cross-naming} representations $\bar{\mathbf{r}}_a^{L,T}$ for each action and predicate, compute the overall mean $\bar{\mathbf{r}}_{\text{mean}}^{L,T}$ across all actions (and separately for predicates), and construct symbolic representations as:
\begin{equation}
    \mathbf{r}_{\text{symbolic}}[a] = \bar{\mathbf{r}}_{\text{mean}}^{L,T} + s \cdot \bar{\mathbf{r}}_{a}^{L,T}
\end{equation}
where $s$ is a mixing scale and $\bar{\mathbf{r}}_{a}^{L,T}$ is the centered cross-naming representation of action $a$.

\paragraph{Experimental Design.} Since the LLM maintained reasonable accuracy even when all actions were replaced with a single vector, we use a comparative approach: \textbf{(1)} Symbolic Patching replaces residual stream activations for action/predicate tokens with corresponding symbolic representations, and \textbf{(2)} Shuffled Patching uses randomly permuted symbolic representations as control. We patch token window $[2000, 4000]$ on all layers up to a selected end layer, then measure accuracy difference $\text{Acc}_{\text{symbolic}} - \text{Acc}_{\text{shuffled}}$. Across all tested mixing scales and end layers, \textbf{matched symbolic representations} consistently outperform shuffled ones (\Cref{fig:patching_scales}, in the appendix), supporting criterion~\textbf{C3}: the cross-naming construction carries information that the shuffled control destroys, even when both are inserted at the same layers and magnitudes. Because both conditions patch the same positions with the same vector construction, the accuracy gap reflects whether the inserted vector is assigned to the correct action or predicate.

\subsection{Negative Steering} \label{sec:steering_negative}

To further test whether refined in-naming representations are functionally used, we conduct negative steering by subtracting converged in-naming representations from LLM activations. Since steering interventions can easily degrade performance, we use shuffled representations as a comparative baseline. Performance degradation beyond the shuffled control demonstrates that the adaptations encode operationally important structural information rather than arbitrary activation patterns.

\paragraph{Experimental Design.} We perform interventions across token window $[2000, 4000]$ on multiple layers, subtracting centered naming representations extracted from the 4k timestamp (chosen because representations are near convergence while still at our window's end, see \Cref{fig:diff_timestamps}). We use shuffled in-naming representations as control rather than Gaussian noise, because shuffling preserves the per-naming activation statistics and only breaks the action--vector correspondence; this is exactly the null we need to rule out for criterion~\textbf{C3}.

\paragraph{Result.} Subtracting refined in-naming representations degrades held-out accuracy by $2.9\%\!\pm\!1.06\%$ (start layer 10, end layer 30) relative to a shuffled control of equal magnitude, and by $2.3\%\!\pm\!0.99\%$ at end layer 20 (full per-layer results in \Cref{app:negative_steering}). Because the shuffled control matches activation magnitude, per-naming statistics, and intervention layers exactly, the gap is not attributable to general intervention disruption. Together with positive steering, negative steering shows the same representation directions from both sides: \textbf{injecting} refined directions can improve held-out accuracy, while \textbf{subtracting} matched directions selectively degrades it.

\subsection{Replacement Steering}
\label{sec:replacement_steering}

The preceding interventions use naming-specific or cross-naming vectors extracted from obfuscated traces. As a stricter causal probe, we test whether the centered clean-domain centroid for an action is itself read by downstream layers as a coherent substitute for the obfuscated representation. At each LLM's default steering layer, we install a forward hook that replaces the residual stream activation at obfuscated phrase positions with the corresponding clean-naming centroid, using positional alignment between obfuscated and clean phrase orderings. We compare against baseline and ablation controls, scoring per-token log-likelihood of each obfuscated trace's own continuation tokens over 30 traces per cell.

On Nemotron-49B-instruct, clean-centroid replacement is less disruptive than zeroing for coherent Mystery BlocksWorld namings, but random-letter naming erases the advantage and Logistics does not show the same substitution effect. We re-ran the probe on Qwen2.5-32B-instruct, Qwen2.5-32B-base, and the two highest-C1 LLMs outside QwQ-32B; across these four LLMs, clean-centroid replacement is consistently more disruptive than zeroing on BlocksWorld, while the two probes are statistically indistinguishable on Logistics. Replacement steering therefore supports a narrow functional reading: \textbf{geometric alignment} of obfuscated and clean centroids does \emph{not} automatically imply \textbf{functional interchangeability}, and the clean centroid acts as a coherent substitute only in the Nemotron BlocksWorld cell. The cross-naming-vs-Gaussian asymmetry of \Cref{sec:steering_positive} remains the cleanest C3 evidence overall.

\section{Conclusion}

We introduced \emph{Fluid Reasoning Representations} as a mechanistic account of how LLMs organize abstract task concepts during extended test-time thinking. Across obfuscated reasoning tasks and multiple open-weight models, action and predicate representations become increasingly tied to their latent roles rather than their surface forms, and interventions on these representations affect downstream reasoning. The evidence suggests that test-time thinking is not only a longer search process: it can also construct \textbf{task-specific symbol meanings} in activation space. Comparisons with base and instruction-tuned models further show that they already exhibit the same dynamic at lower magnitude, so extended test-time thinking amplifies and operationally exploits a \textbf{pre-existing representational substrate} rather than creating it from scratch. FRRs therefore offer a concrete view on the internal dynamics behind thinking behavior at test time in complex problem-solving.

\section*{Limitations}

Our evidence is concentrated on structured reasoning settings where action and predicate phrases can be aligned across obfuscated and clean variants. The interventions are controlled probes over fixed token windows and should be interpreted as evidence about representational dynamics rather than as general-purpose accuracy interventions. Cross-domain analyses include Logistics and GSM8K-style renaming, but the strongest functional evidence remains in the BlocksWorld setting; broader domains and model families are left for future work.

\bibliography{custom}

\appendix

\section{Potential Risks}
\label{app:responsible_checklist}

\noindent This work studies controlled representational interventions for mechanistic analysis, not deployment-oriented model-control methods. The main potential risk is that steering-style interventions could be over-generalized as practical model-control techniques. We mitigate this by using synthetic obfuscated reasoning tasks, reporting aggregate behavioral effects, and framing the interventions as diagnostic probes of whether task representations matter for later reasoning.

\paragraph{Scientific artifacts and licenses.} We use public or cited scientific artifacts, including BlocksWorld and PlanBench task formats, GSM8K-style examples, public model checkpoints or APIs, and inference and interpretability software. PlanBench/LLMs-Planning, GSM8K, TransformerLens, and NNSight are MIT-licensed; vLLM, QwQ-32B, Qwen2.5-32B base/instruct, and Seed-OSS-36B base/instruct are Apache-2.0-licensed. DeepSeek-R1-Distill-Qwen-32B and DeepSeek-R1-Distill-Llama-70B are released under the MIT License, with the model cards also noting Apache-2.0 terms for the Qwen-derived model and Llama 3.3 terms for the Llama-derived model. Llama-3.3-70B-Instruct is used under the Llama 3.3 Community License, Nemotron-49B-instruct under the NVIDIA Open Model License with Llama 3.3 terms, and GPT-4.1 through the OpenAI API terms. The corresponding creators are cited in the main text, related work, and implementation details. We use all artifacts only for academic research evaluation and mechanistic analysis, strictly follow their intended use and license terms, and do not redistribute model weights or package the artifacts as deployable systems.

\paragraph{Data content and documentation.} The data consists of synthetic planning states, symbolic action and predicate vocabularies, GSM8K-style math-word-problem variants, model-generated traces, and aggregate evaluation results. The tasks are in English and synthetic symbolic domains; they do not represent demographic groups and do not contain private user data or newly collected human-subject data. The generated task text and obfuscation vocabularies use synthetic symbols rather than real-person names, and we checked them by construction and manual inspection for identifying or offensive content.

\paragraph{Data statistics.} The main text and appendix report the relevant dataset and evaluation statistics. Mystery BlocksWorld uses 300 four-block puzzles and 15 primary obfuscation namings for QwQ-32B analyses, with naming 3 excluded from representation analyses because it bypasses the intended adaptation process. Cross-domain C1 analyses use Logistics and GSM8K-Renamed variants with the number of namings and traces reported in \Cref{tab:phase1_c1}. Steering vectors are constructed from 40 correctly solved puzzles and evaluated on 100 held-out four-block puzzles per naming; additional hyperparameters and splits are reported in \Cref{app:hyperparameters}.

\paragraph{Computational experiments.} All computational experiments are inference-only: we do not train or fine-tune LLMs. Model families and sizes are reported in the main text and model names where applicable. \Cref{app:hyperparameters,app:implementation_details} report representation extraction windows, timestamps, layers, steering windows, steering scales, decoding settings, package versions, and implementation details. Exact total GPU-hour accounting is not used to support any claim in the paper; the inference and intervention stack is vLLM v0.7.3 with PyTorch forward hooks.

\paragraph{Descriptive statistics.} We report accuracies, token counts, cosine similarities, mean steering improvements, standard errors, and paired significance tests where appropriate. Figures and tables specify whether values are means, held-out accuracies, cosine similarities, or best per-family/cell measurements.

\paragraph{Human subjects.} This work does not use human subjects, crowdworkers, annotators, or newly collected human-subject data.

\section{AI Assistant Use}
\label{app:ai_assistant_use}

\noindent AI assistants were used for language polishing, LaTeX editing, and code-writing support. The authors checked the final manuscript, experimental claims, numerical results, and citations, and AI assistants are not listed as authors.

\section{BlocksWorld Prompt Example}
\label{app:prompt_blocksworld}

\lstset{basicstyle=\footnotesize\ttfamily,breaklines=true,frame=none,xleftmargin=0pt,aboveskip=0pt,belowskip=0pt}
\begin{promptbox}{BlocksWorld system prompt}
\begin{lstlisting}
I am playing with a set of blocks where I need to arrange the blocks into stacks. Here are the actions I can do

Pick up a block
Unstack a block from on top of another block
Put down a block
Stack a block on top of another block

I have the following restrictions on my actions:
I can only pick up or unstack one block at a time.
I can only pick up or unstack a block if my hand is empty.
I can only pick up a block if the block is on the table and the block is clear.
A block is clear if the block has no other blocks on top of it and if the block is not picked up.
I can only unstack a block from on top of another block if the block I am unstacking was really on top of the other block.
I can only unstack a block from on top of another block if the block I am unstacking is clear.
Once I pick up or unstack a block, I am holding the block.
I can only put down a block that I am holding.
I can only stack a block on top of another block if I am holding the block being stacked.
I can only stack a block on top of another block if the block onto which I am stacking the block is clear.
Once I put down or stack a block, my hand becomes empty.
Once you stack a block on top of a second block, the second block is no longer clear.

Here is an example problem:

[STATEMENT]
As initial conditions I have that, Block B is clear, Block C is clear, the hand is empty, Block C is on top of Block A, Block A is on the table, Block B is on the table.
My goal is to have that Block A is on top of Block C and Block B is on top of Block A

My plan is as follows:

[PLAN]
unstack Block C from on top of Block A
put down Block C
pick up Block A
stack Block A on top of Block C
pick up Block B
stack Block B on top of Block A
[PLAN END]
\end{lstlisting}
\end{promptbox}

\section{Mystery Prompt Example}
\label{app:prompt}

\begin{promptbox}{Mystery BlocksWorld system prompt (naming 1)}
\begin{lstlisting}
I am playing with a set of objects. Here are the actions I can do:
   Attack object
   Feast object from another object
   Succumb object
   Overcome object from another object

I have the following restrictions on my actions:
    To perform Attack action, the following facts need to be true: Province object, Planet object, Harmony.
    Once Attack action is performed the following facts will be true: Pain object.
    Once Attack action is performed the following facts will be false: Province object, Planet object, Harmony.
    To perform Succumb action, the following facts need to be true: Pain object.
    Once Succumb action is performed the following facts will be true: Province object, Planet object, Harmony.
    Once Succumb action is performed the following facts will be false: Pain object.
    To perform Overcome action, the following needs to be true: Province other object, Pain object.
    Once Overcome action is performed the following will be true: Harmony, Province object, Object Craves other object.
    Once Overcome action is performed the following will be false: Province other object, Pain object.
    To perform Feast action, the following needs to be true: Object Craves other object, Province object, Harmony.
    Once Feast action is performed the following will be true: Pain object, Province other object.
    Once Feast action is performed the following will be false:, Object Craves other object, Province object, Harmony.

Here is an example problem:
[STATEMENT]
As initial conditions I have that, province Block B, province Block C, harmony, Block C craves Block A, planet Block A, planet Block B.
My goal is to have that Block A craves Block C and Block B craves Block A.
My plan is as follows:
[PLAN]
feast Block C from Block A
succumb Block C
attack Block A
overcome Block A from Block C
attack Block B
overcome Block B from Block A
[PLAN END]
\end{lstlisting}
\end{promptbox}

\section{Behavior Analysis}
\label{app:behavior}
Through manual inspection of DeepSeek-R1-Distill-Qwen-32B and QwQ-32B traces, we identified recurring behavioral patterns in Mystery BlocksWorld solving. LLMs begin with \textbf{comparative analysis}, examining initial and goal states to identify conflicting predicates. They then alternate between \textbf{recursive search} (working backwards from goals to identify required actions) and \textbf{exploration} (experimenting with actions to discover achievable states). These exploratory behaviors occupy the first half of reasoning traces. The second phase involves \textbf{plan formulation}, where LLMs construct action sequences and verify validity, iteratively rebuilding when conflicts arise. The final phase consists of \textbf{plan verification}, where LLMs validate solutions before committing to answers.

\section{Layer-wise PCA}
\label{app:pca}
\begin{figure*}[!htbp]
    \centering
    \begin{subfigure}[t]{\textwidth}
        \centering
        \includegraphics[width=\linewidth]{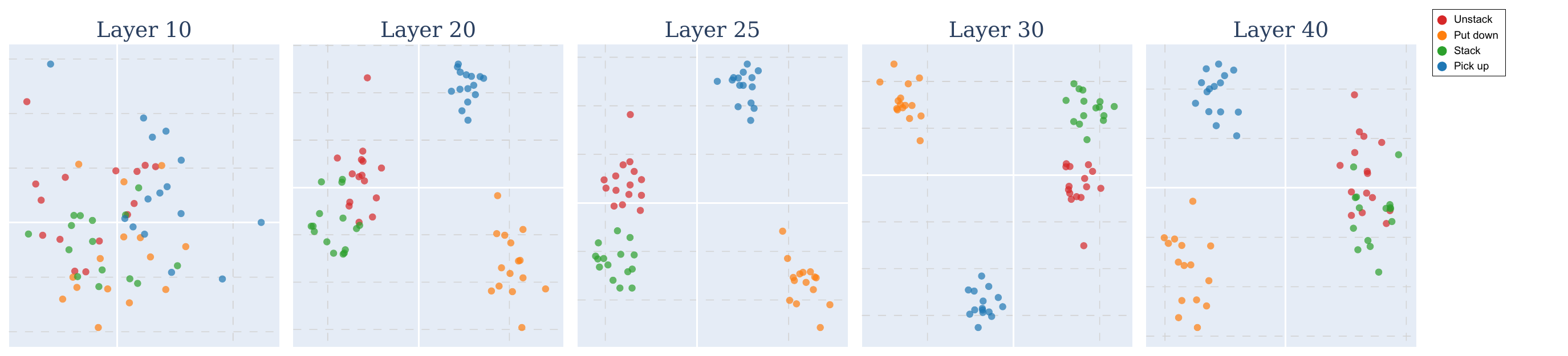}
        \caption{Nemotron-49B-instruct}
    \end{subfigure}

    \vspace{0.5em}

    \begin{subfigure}[t]{\textwidth}
        \centering
        \includegraphics[width=\linewidth]{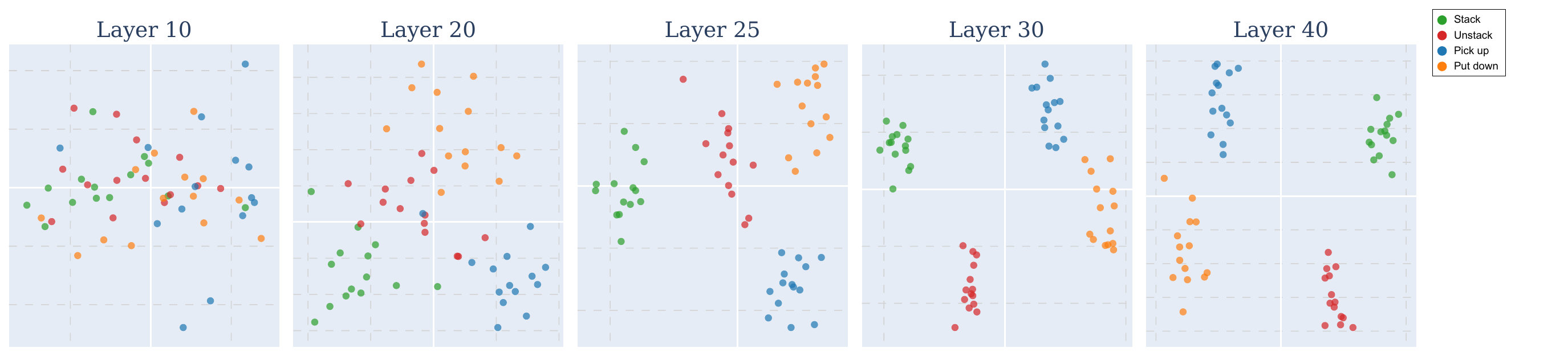}
        \caption{Seed-OSS-36B-instruct}
    \end{subfigure}

    \caption{Layer-wise action PCA at 7k tokens.}
    \label{fig:pca_7k_all}
\end{figure*}

\cref{fig:pca_7k_all} contains layer-wise PCA for action representations of different LLMs. The clustering patterns become more pronounced in deeper layers, with clear separation between action types emerging around layers 20--30.

\begin{figure*}[!htbp]
    \centering
    \begin{subfigure}[t]{\textwidth}
        \centering
        \includegraphics[width=\linewidth]{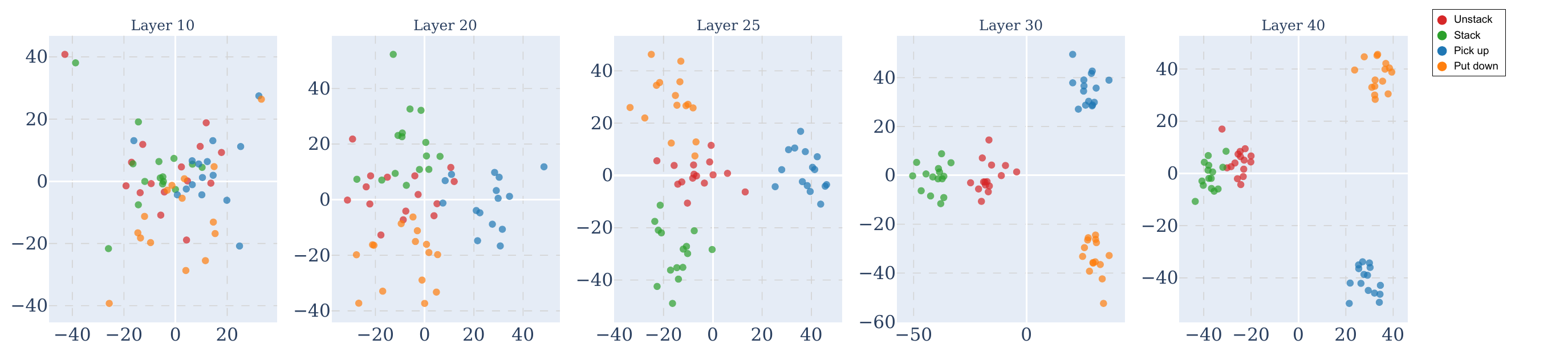}
        \caption{QwQ}
        \label{fig:pca_7k_qwq}
    \end{subfigure}

    \vspace{0.5em}

    \begin{subfigure}[t]{\textwidth}
        \centering
        \includegraphics[width=\linewidth]{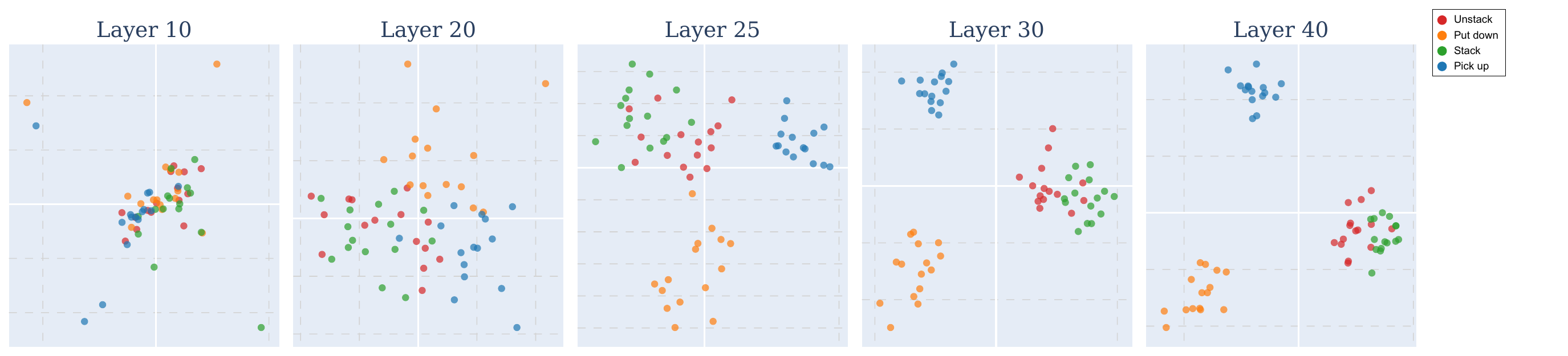}
        \caption{DeepSeek-R1-Distill-Qwen-32B}
    \end{subfigure}

    \caption{Layer-wise action PCA at 7k tokens.}
\end{figure*}

\section{Cross-LLM Similarity Analysis}
\label{app:cross_model_similarities}

\begin{figure*}[!htbp]
    \centering
    \begin{subfigure}[t]{0.48\textwidth}
        \centering
        \includegraphics[width=\linewidth]{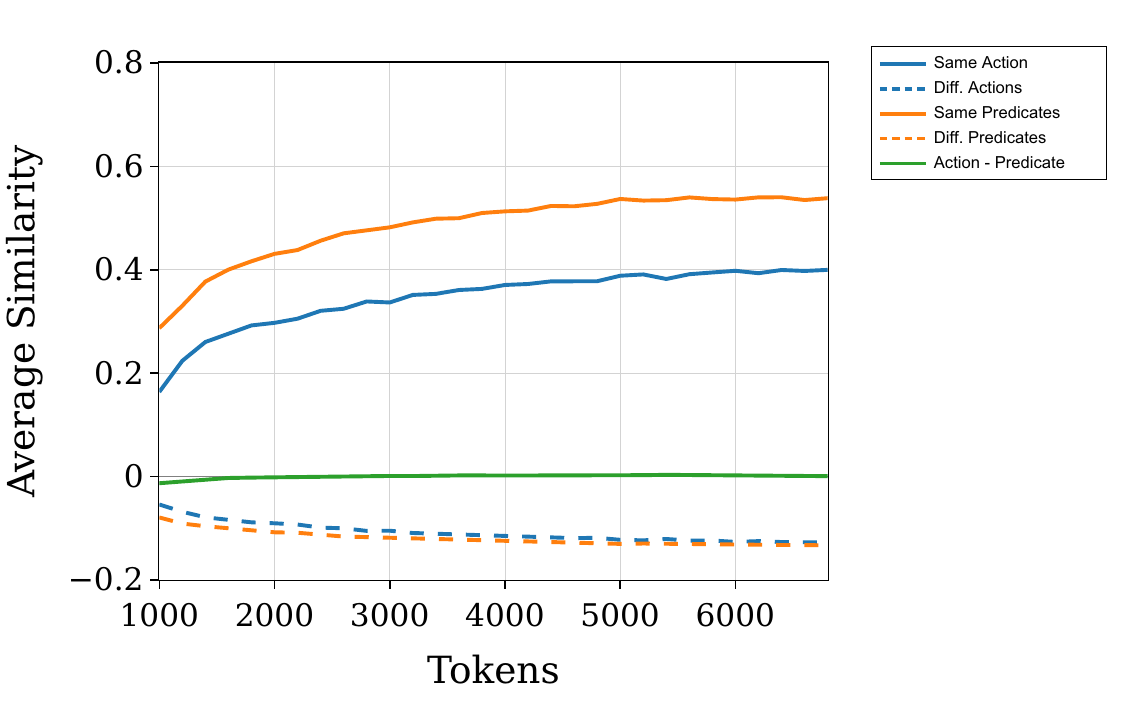}
        \caption{DeepSeek-R1-Distill-Qwen-32B.}
        \label{fig:sims_avg_ds}
    \end{subfigure}
    \hfill
    \begin{subfigure}[t]{0.48\textwidth}
        \centering
        \includegraphics[width=\linewidth]{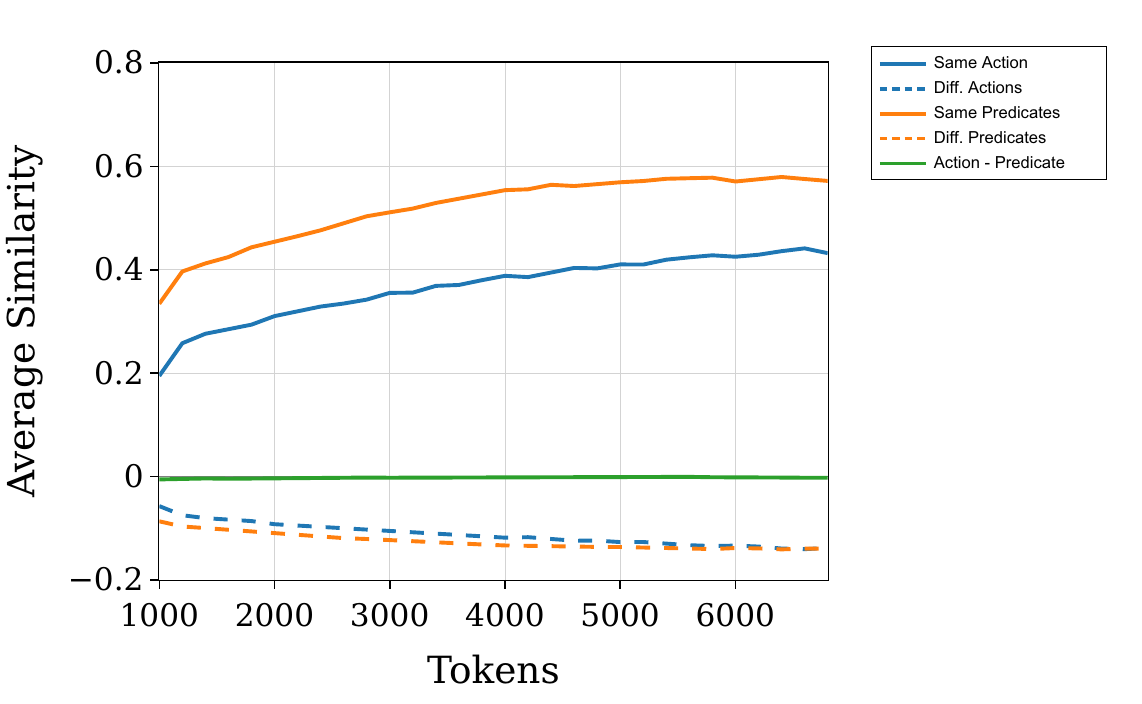}
        \caption{Nemotron-49B-instruct.}
        \label{fig:sims_avg_nemo}
    \end{subfigure}
    \caption{Similarities with cross-naming average representations.}
\end{figure*}

\begin{figure*}[!htbp]
    \centering
    \begin{subfigure}[t]{0.48\textwidth}
        \centering
        \includegraphics[width=\linewidth]{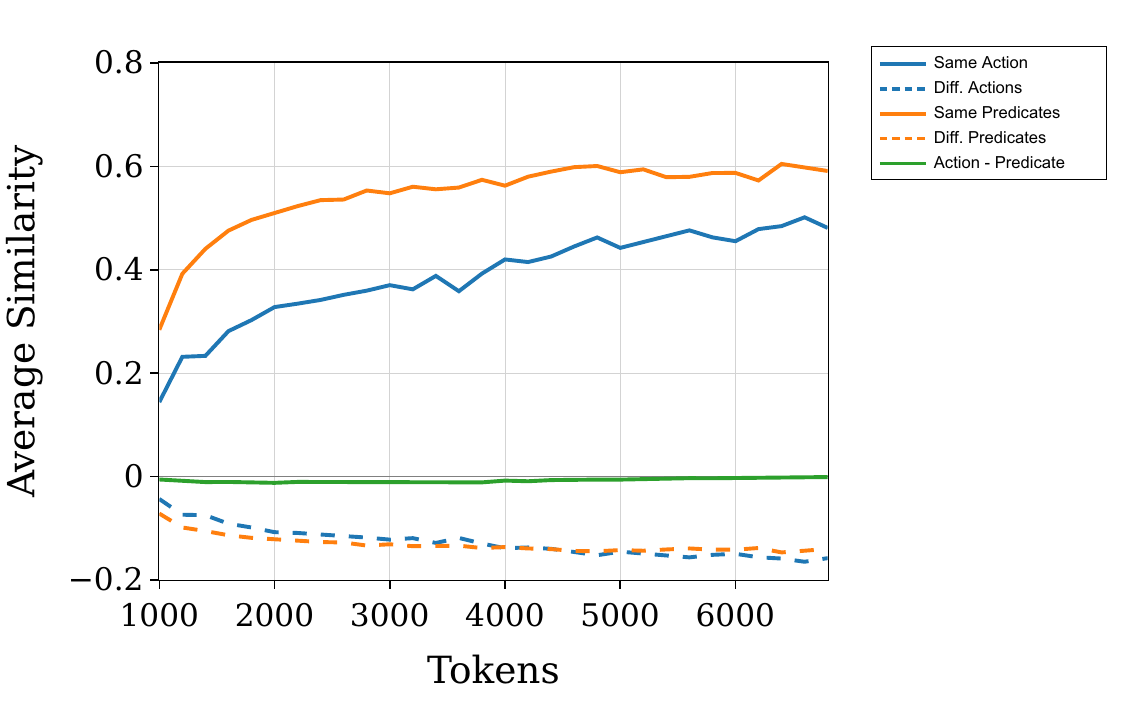}
        \caption{Seed-OSS-36B-instruct.}
        \label{fig:sims_avg_seed}
    \end{subfigure}
    \hfill
    \begin{subfigure}[t]{0.48\textwidth}
        \centering
        \includegraphics[width=\linewidth]{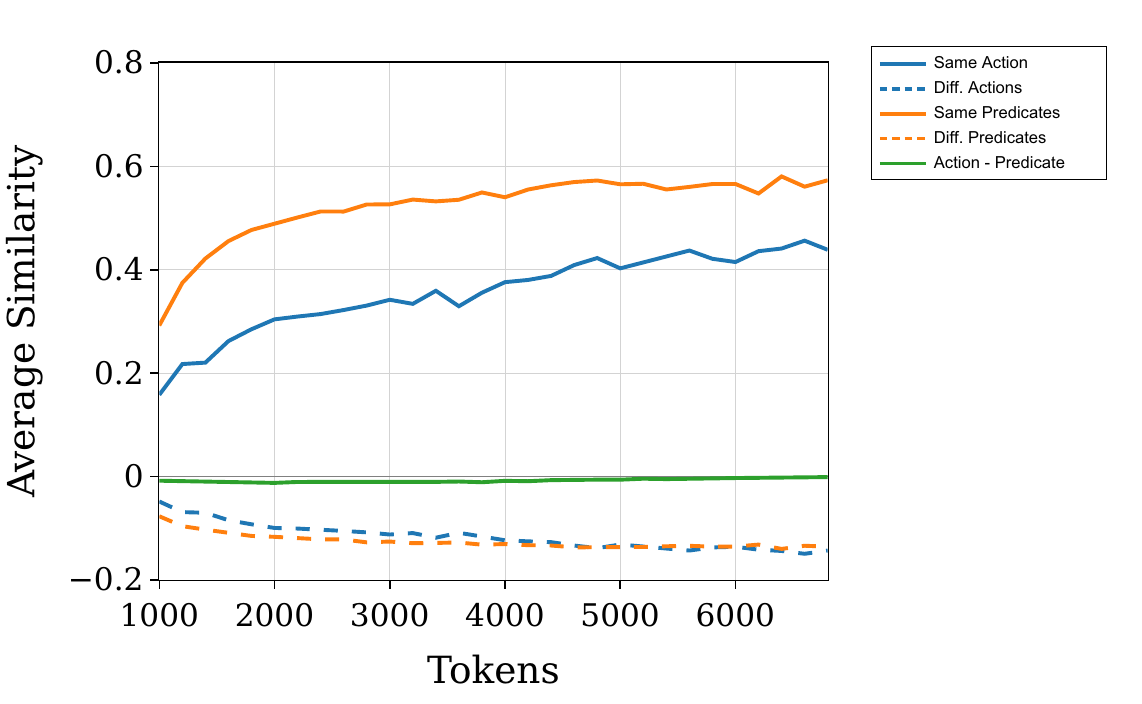}
        \caption{Seed-OSS-36B-base \citep{seed2025seed-oss}.}
        \label{fig:sims_avg_seed_base}
    \end{subfigure}
    \caption{Similarities with cross-naming average representations.}
\end{figure*}

To validate that representational convergence is not unique to QwQ-32B, we analyzed action and predicate representations across multiple LLMs. \Cref{fig:sims_avg_ds,fig:sims_avg_nemo,fig:sims_avg_seed,fig:sims_avg_seed_base} show how centered action and predicate representations from different timestamps converge toward cross-naming average representations extracted at 7k tokens, analogous to the analysis in \Cref{sec:rep_avg}.

\section{Hyperparameters and Experimental Configuration}
\label{app:hyperparameters}

This section summarizes the hyperparameters and experimental settings used throughout our analysis.

\subsection{Representation Collection}
\label{app:hyperparams_representation}

\begin{itemize}
    \item \textbf{Phrase-matching window}: 100 tokens before each timestamp.
    \item \textbf{Extraction stride and reported timestamps}: Hidden states are sampled every 200 tokens; reported analyses use 2k, 4k, 7k, and 10k.
    \item \textbf{Batch size} ($b$): Number of reasoning traces per batch for representation extraction. We use 40 correctly solved traces for steering-vector construction and reserve 100 held-out puzzles for steering evaluation.
    \item \textbf{Layers analyzed}: All layers from 0 to LLM depth
    \item \textbf{Number of puzzles for representation collection}: 40 correctly solved puzzles for steering vectors, full dataset for general analysis
    \item \textbf{Layer selection for representation analysis}: We selected layer 40 because action representations have largely separated by this point and change only modestly in later layers. We also performed the analysis at layer 30 and observed no qualitatively different conclusions.
\end{itemize}

\subsection{Positive Steering}
\label{app:hyperparams_positive}

\begin{itemize}
    \item \textbf{Steering scale} ($s$): $\frac{2}{3}$ (selected after sweep, see \Cref{fig:20_sweep})
    \item \textbf{Steering window}: $[t_{\text{start}}, t_{\text{end}}) = [1500, 2500]$ tokens
    \item \textbf{Layers tested}: 1, 5, 10, 20, 30, 40, 50, 60
    \item \textbf{Intervention dataset}: 100 held-out 4-block problem rollouts
    \item \textbf{Steering vector extraction timestamp}: 7k tokens (for layer-wise analysis)
\end{itemize}

\subsection{Symbolic Patching}
\label{app:hyperparams_patching}

\begin{itemize}
    \item \textbf{Patching window}: $[2000, 4000]$ tokens
    \item \textbf{End layers tested}: Multiple layers up to selected end layer
    \item \textbf{Scaling factors} ($s$): $\{10, 20\}$
    \item \textbf{Symbolic representation construction}: $\mathbf{r}_{\text{symbolic}}[a] = \mathbf{r}_{\text{mean}} + s \cdot \mathbf{r}_{a}$
    \item \textbf{Control condition}: Shuffled symbolic representations (random permutation)
\end{itemize}

\begin{figure}[!htbp]
    \centering
    \includegraphics[width=\columnwidth]{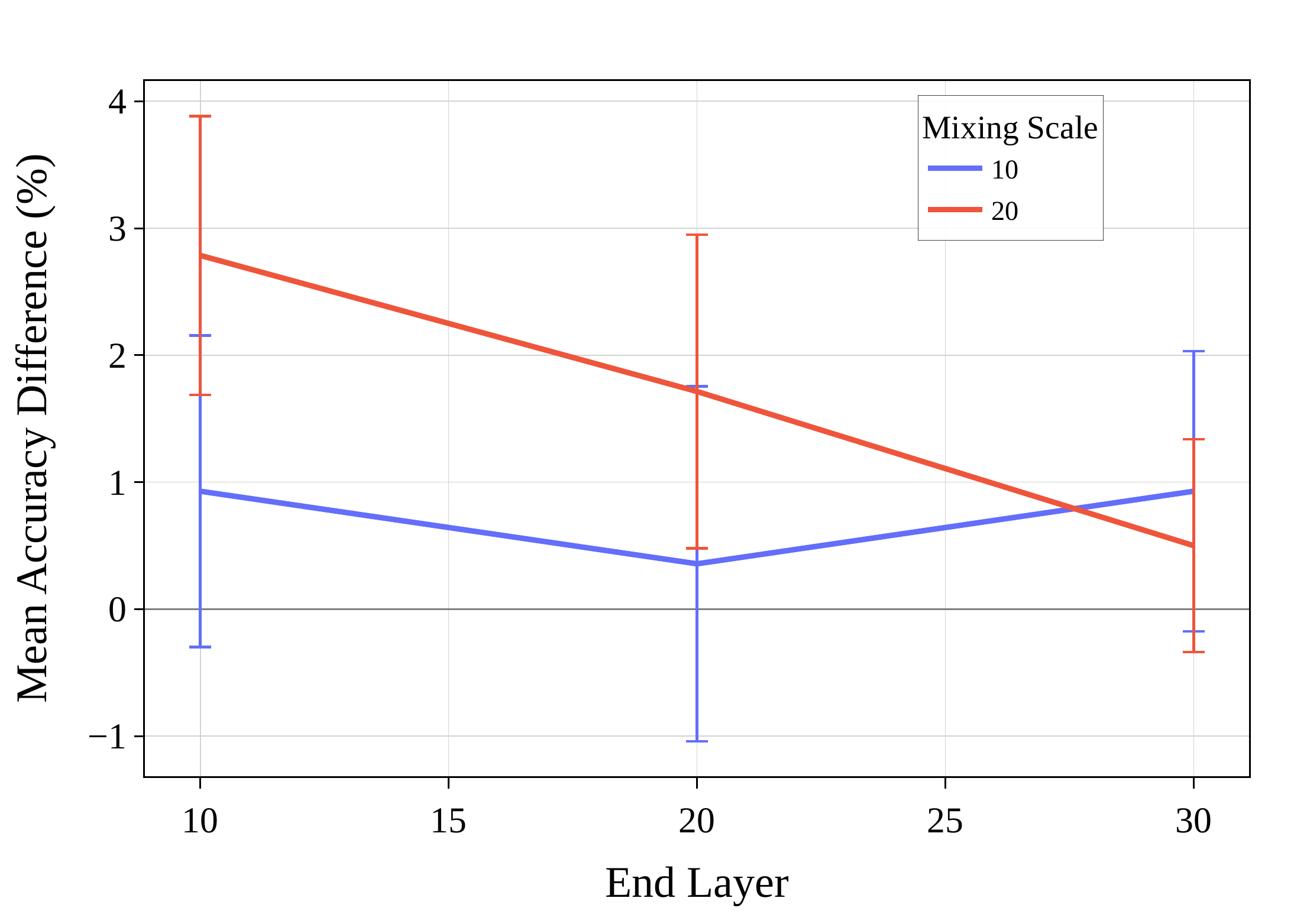}
    \caption{\textbf{Symbolic patching.} Mean $\text{Acc}_{\text{symbolic}} - \text{Acc}_{\text{shuffled}}$ across scales and end layers; error bars show standard error.}
    \label{fig:patching_scales}
\end{figure}

\subsection{Negative Steering}
\label{app:hyperparams_negative}

\begin{itemize}
    \item \textbf{Intervention window}: $[2000, 4000]$ tokens
    \item \textbf{Representation extraction timestamp}: 4k tokens
    \item \textbf{Start layer}: 10
    \item \textbf{End layers tested}: 20, 30
    \item \textbf{Control condition}: Shuffled centered naming representations
\end{itemize}

\subsection{LLM Inference}
\label{app:hyperparams_inference}

\begin{itemize}
    \item \textbf{Decoding strategy}: Greedy decoding
    \item \textbf{Maximum sequence length}: 24,576 tokens
    \item \textbf{Temperature}: 0 (greedy)
    \item \textbf{Implementation}: vLLM v0.7.3 with PyTorch forward hooks
\end{itemize}

\subsection{Dataset Configuration}
\label{app:hyperparams_dataset}

\begin{itemize}
    \item \textbf{Number of puzzles}: 300 four-block BlocksWorld puzzles
    \item \textbf{Number of mystery namings}: 15 (primary experiments), 20 (including additional variants)
    \item \textbf{Naming 3 exclusion}: Excluded from representational analyses (recognized as BlocksWorld)
    \item \textbf{Train/test split}: 40 correctly solved puzzles for steering vector extraction, 100 held-out puzzles for steering evaluation
\end{itemize}

\section{Negative Steering}
\label{app:negative_steering}

To further validate the causal role of FRRs, we conduct an ablation experiment testing whether disrupting representational adaptations decreases accuracy. Since steering interventions can easily degrade performance through general disruption rather than targeted ablation, we use a comparative approach.

We perform interventions across token window $[2000, 4000]$ on multiple layers, subtracting centered in-naming representations extracted from the 4k timestamp. We use shuffled in-naming representations as the control because they preserve activation statistics while breaking action-vector correspondence. Starting at layer 10, end layer 20 degrades accuracy by $2.3\%\!\pm\!0.99\%$ relative to the shuffled control, and end layer 30 degrades accuracy by $2.9\%\!\pm\!1.06\%$.

\section{Mystery BlocksWorld Naming Variants} \label{app:variants}

\begin{table*}[t]
\centering
\caption{Action Mappings Across Mystery Namings}
\small
\setlength{\tabcolsep}{6pt}
\begin{tabular}{|l|l|l|l|l|}
\hline
\textbf{Naming} & \textbf{pick up} & \textbf{put down} & \textbf{stack} & \textbf{unstack} \\
\hline
Mystery 1 & attack & succumb & overcome & feast \\
Mystery 2 & illuminate & silence & distill & divest \\
Mystery 3 & tltezi & jchntg & deesdu & xavirm \\
Mystery 4 & swim & fire & deduct & respond \\
Mystery 5 & whisper & calculate & orbit & navigate \\
Mystery 6 & decode & hibernate & thunder & quench \\
Mystery 7 & explore & ripen & weave & bloom \\
Mystery 8 & harvest & ignite & carve & suspend \\
Mystery 9 & construct & demolish & reinforce & collapse \\
Mystery 10 & plant & harvest & nurture & prune \\
Mystery 11 & prosecute & acquit & testify & appeal \\
Mystery 12 & broadcast & receive & encrypt & decode \\
Mystery 13 & whisper & banish & entangle & unmask \\
Mystery 14 & question & resolve & interweave & liberate \\
Mystery 15 & summon & dismiss & fold & unravel \\
\hline
\hline
\multicolumn{5}{|c|}{\textbf{Additional Naming Variants}} \\
\hline
\hline
Mystery 16 & open & close & connect & disconnect \\
Mystery 17 & chop & serve & season & taste \\
Mystery 18 & release & grasp & separate & combine \\
Mystery 19 & transcend & sublimate & actualize & deconstruct \\
Mystery 20 & flixate & grample & chonder & sprill \\
\hline
\end{tabular}
\end{table*}

\begin{table*}[t]
\centering
\caption{Predicate Mappings Across Mystery Namings}
\small
\setlength{\tabcolsep}{5pt}
\begin{tabular}{|l|l|l|l|l|l|}
\hline
\textbf{Naming} & \textbf{ontable} & \textbf{clear} & \textbf{handempty} & \textbf{holding} & \textbf{on} \\
\hline
Mystery 1 & planet & province & harmony & craves & pain \\
Mystery 2 & aura & essence & nexus & harmonizes & pulse \\
Mystery 3 & oxtslo & adohre & jqlyol & gszswg & ivbmyg \\
Mystery 4 & fever & marble & craving & mines & shadow \\
Mystery 5 & crystal & fountain & autumn & illuminates & legend \\
Mystery 6 & prism & hollow & zenith & echoes & emblem \\
Mystery 7 & fossil & dialect & equinox & fractures & symphony \\
Mystery 8 & nebula & labyrinth & mirage & captivates & cascade \\
Mystery 9 & eclipse & vintage & paradox & resonates & twilight \\
Mystery 10 & crystal & puzzle & vortex & whispers & cipher \\
Mystery 11 & nebula & molecule & anthem & silhouettes & voltage \\
Mystery 12 & horizon & compass & solstice & orbits & quantum \\
Mystery 13 & tethered & unburdened & hollow & shrouds & consuming \\
Mystery 14 & echoing & sovereign & potential & obscures & contemplating \\
Mystery 15 & suspended & timeless & interval & transcends & enveloping \\
\hline
\hline
\multicolumn{6}{|c|}{\textbf{Additional Naming Variants}} \\
\hline
\hline
Mystery 16 & paired & single & balanced & matches & mirrors \\
Mystery 17 & plated & fresh & kitchen & simmering & marinated \\
Mystery 18 & floating & occupied & crowded & repels & avoids \\
Mystery 19 & phenomenal & unmediated & dialectical & instantiates & necessitates \\
Mystery 20 & morkled & thristy & plimmish & vexates & quorbles \\
\hline
\end{tabular}
\end{table*}

\section{Mystery Performance Analysis}
\label{app:mystery_performance}

\begin{table*}[t]
\centering
\caption{Mystery BlocksWorld performance and maximum steering improvement by naming. ``--'' indicates variants not used in the steering sweep.}
\label{tab:mystery_performance_full}
\small
\setlength{\tabcolsep}{4pt}
\renewcommand{\arraystretch}{0.95}
\begin{tabular}{|l|c|c|c|p{6cm}|}
\hline
\textbf{Naming} & \textbf{Base} & \textbf{In-Naming} & \textbf{Cross-Naming} & \textbf{Semantic Description} \\
\textbf{Variant} & \textbf{Acc.} & \textbf{Steering} & \textbf{Steering} & \\
\hline
Mystery 1 & 0.33 & +0.10 & +0.11 & Mixed violent/consumption metaphors \\
Mystery 2 & 0.47 & +0.05 & +0.05 & Abstract mystical/spiritual terms \\
Mystery 3 & 0.65 & -- & -- & Random strings \\
Mystery 4 & 0.25 & +0.03 & +0.03 & Mixed physical actions \\
Mystery 5 & 0.24 & -0.01 & +0.01 & Communication/navigation metaphors \\
Mystery 6 & 0.26 & +0.05 & +0.07 & Technical/elemental operations \\
Mystery 7 & 0.19 & +0.02 & +0.03 & Nature/growth cycle \\
Mystery 8 & 0.11 & +0.02 & +0.04 & Agriculture/crafting metaphors \\
Mystery 9 & 0.25 & +0.02 & +0.01 & Construction/destruction cycle \\
Mystery 10 & 0.05 & +0.09 & +0.05 & Coherent gardening domain \\
Mystery 11 & 0.14 & +0.00 & +0.02 & Legal proceedings domain \\
Mystery 12 & 0.16 & +0.06 & +0.02 & Communication technology \\
Mystery 13 & 0.48 & +0.06 & +0.06 & Dark mystical operations \\
Mystery 14 & 0.24 & +0.02 & +0.02 & Abstract philosophical inquiry \\
Mystery 15 & 0.34 & +0.04 & +0.04 & Mystical summoning/manipulation \\
\hline
\hline
\multicolumn{5}{|c|}{\textbf{Additional Variants}} \\
\hline
\hline
Mystery 16 & 0.05 & -- & -- & Reversible operations (open/close) \\
Mystery 17 & 0.27 & -- & -- & Coherent cooking domain \\
Mystery 18 & 0.07 & -- & -- & Physical manipulation verbs \\
Mystery 19 & 0.33 & -- & -- & Abstract philosophical concepts \\
Mystery 20 & 0.29 & -- & -- & Complete nonsense words \\
\hline
\end{tabular}
\end{table*}

\Cref{tab:mystery_performance_full} reveals several patterns supporting our hypothesis that semantic coherence impedes abstraction. Namings with coherent alternative domains (Mystery 10: gardening, Mystery 11: legal proceedings, Mystery 16: reversible operations) achieve the lowest base accuracies, while abstract or semantically incoherent combinations (Mystery 2, Mystery 13) enable superior performance.

The steering improvement data shows notable heterogeneity across namings. The maximum improvements reported here represent the best performance across layers 20, 30, 40, and 50, as different namings exhibit optimal responsiveness at different depths. Some namings (Mystery 5, Mystery 11) show minimal or no improvement from in-naming steering, while others (Mystery 1, Mystery 10) demonstrate substantial gains. This suggests that certain semantic structures are more amenable to representational refinement than others.

\section{Implementation Details}
\label{app:implementation_details}

\subsection{Steering engine}
\label{app:engine}
We implement steering using PyTorch forward hooks on top of vLLM v0.7.3 \citep{kwon2023efficient}, which provides substantial performance improvements, reducing experiment runtimes from several hours to tens of minutes compared to alternatives like TransformerLens \citep{nanda2022transformerlens} or NNSight \citep{fiottokaufman2024nnsightndifdemocratizingaccess}. However, this approach has tradeoffs: we must be mindful of cache recomputations, since they recompute representations without steering interventions, and vLLM's optimizations introduce some numerical instability during extended reasoning traces. To address this instability, we run experiments across multiple naming variants. All experiments use greedy decoding with maximum sequence length of 24,576 tokens.

\subsection{Hyperparameters}
\label{app:hyperparams}

\begin{figure*}[!htbp]
    \centering
    \begin{subfigure}[t]{0.48\textwidth}
        \centering
        \includegraphics[width=\linewidth]{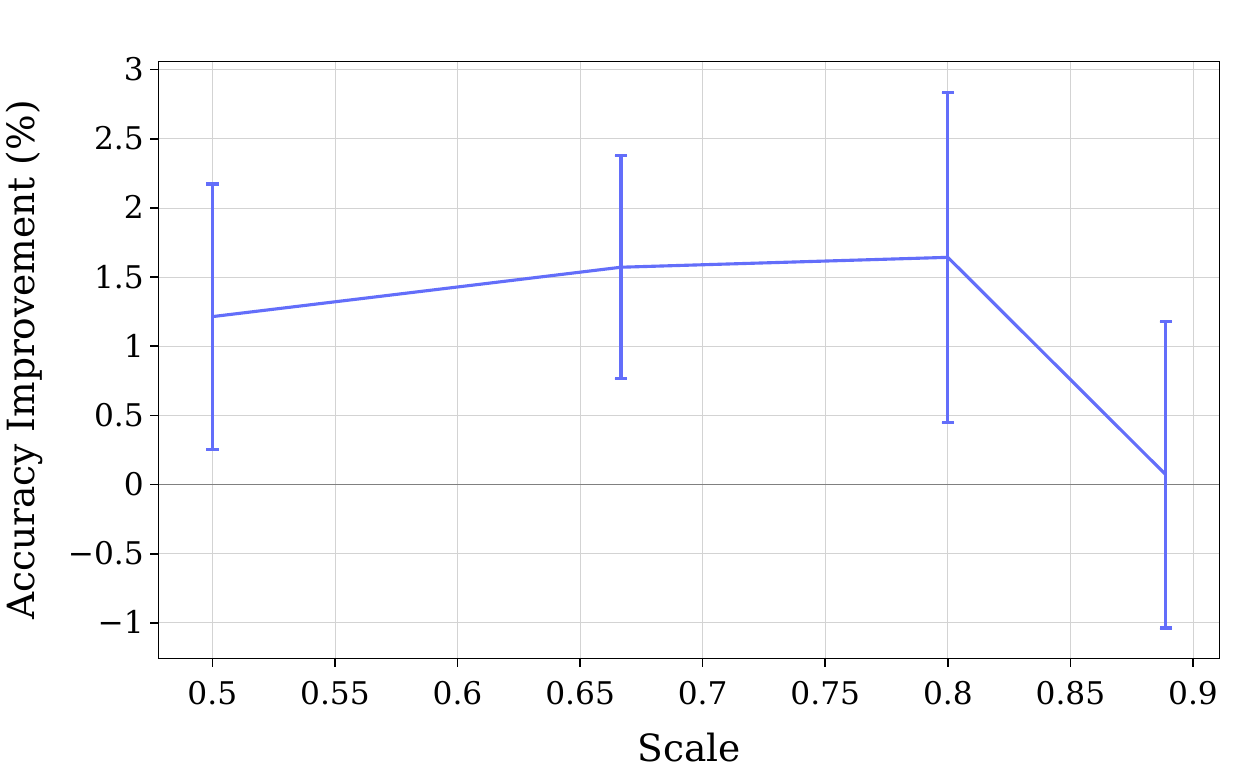}
        \caption{Layer-20 scale sweep.}
        \label{fig:20_sweep}
    \end{subfigure}
    \hfill
    \begin{subfigure}[t]{0.48\textwidth}
        \centering
        \includegraphics[width=\linewidth]{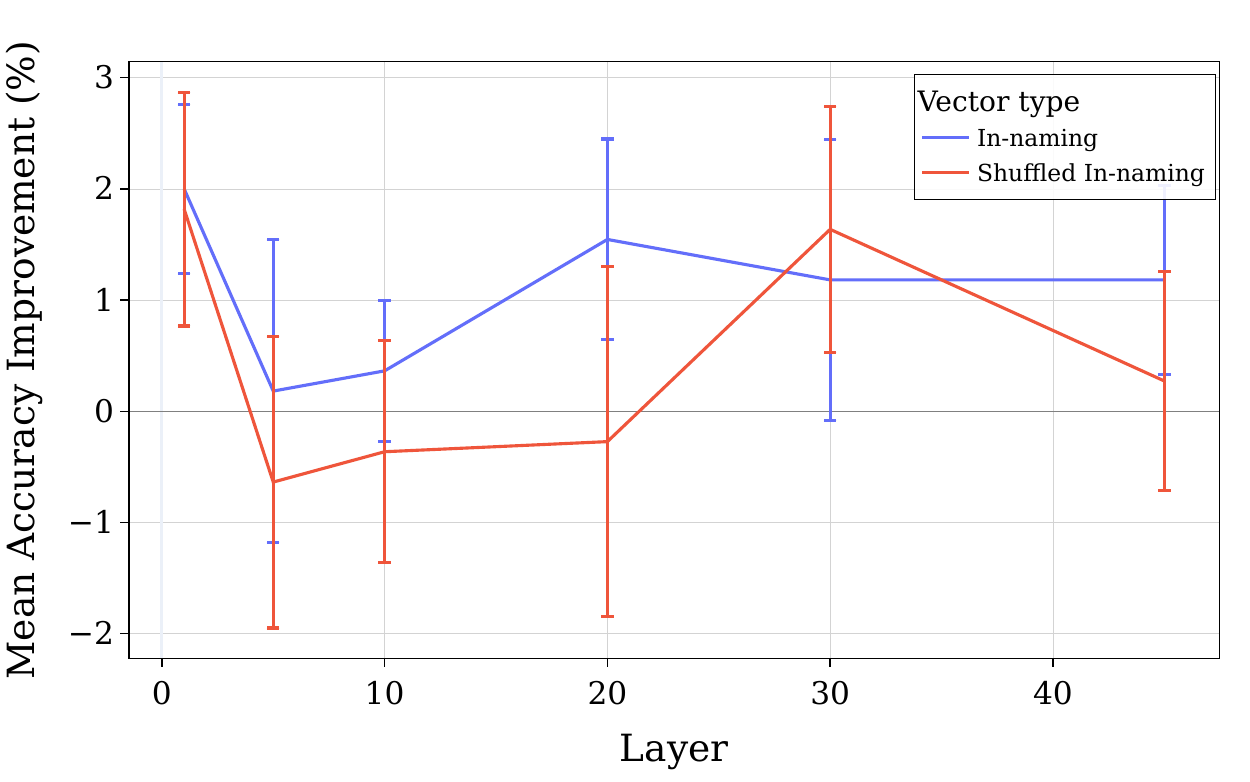}
        \caption{Shuffled in-naming control.}
        \label{fig:shuffled_layers}
    \end{subfigure}
    \caption{Additional steering controls and sweeps.}
\end{figure*}

We perform positive steering as described in \Cref{sec:steering_positive} on layer 20 using \textbf{in-naming} representations to determine the optimal steering scale. \Cref{fig:20_sweep} shows that scales $\frac{2}{3}$ and $\frac{4}{5}$ have similar effects. While improvement from $\frac{4}{5}$ is slightly higher, we chose $\frac{2}{3}$ since it has a more stable effect on all namings.

\section{Shuffled In-Naming Steering Controls}
\label{app:shuffled_steering}

\noindent\textbf{Setup.} In addition to positive steering (\Cref{sec:steering_positive}), we tested a \emph{shuffled in-naming} control. For each naming, we applied a single consistent permutation of centered in-naming vectors across actions/predicates (e.g., \texttt{pick up}$\to$\texttt{stack}, \texttt{stack}$\to$\texttt{put down}, etc.). This preserves per-naming distributional statistics while breaking the action--representation alignment. Interventions used the same window $[1500, 2500]$, scale $s=\tfrac{2}{3}$, and norm-preserving update rule as before, run on a reduced subset of layers and namings.

\noindent\textbf{Findings and interpretation.} \Cref{fig:shuffled_layers} shows a three-phase trend: (i) early layers ($\!\le\!5$) improve relative to baseline, consistent with disruption of surface semantics; (ii) middle layers ($\sim$5--20) degrade performance, likely breaking emerging abstractions; and (iii) later layers ($\ge$30) again show improvements, comparable to unshuffled \emph{in-naming}. Early/mid results align with our story: disruption helps initially, but permutations harm once action-specific representations form. Late-layer gains suggest action vectors contain shared structural components, so even mismatched but in-manifold vectors can occasionally assist; the runs therefore serve as a targeted control for action-vector correspondence under the same steering rule.

\section{Statistical Analysis of Steering Effects}
\label{app:steering_statistics}
\label{app:statistical_analysis}

We conducted statistical tests to validate that steering with refined representations improves accuracy over baseline. Our analysis uses one-sample t-tests treating each mystery naming as an independent observation.

\subsection{Test Methodology}

\noindent\textbf{Data Structure.} Our experiments evaluate steering across 14 mystery namings (excluding naming 3) with approximately 100 puzzles per naming (indices 200--300). Each mystery naming provides accuracy measurements under multiple conditions: baseline (no steering), in-naming steering, cross-naming steering, and random Gaussian steering at various layers.

\noindent\textbf{Statistical Test.} We employ one-sample t-tests to assess whether mean accuracy improvements across mystery namings significantly exceed zero. Each mystery naming serves as an independent observation, with the null hypothesis $H_0: \mu_{\text{improvement}} = 0$. We use one-tailed tests since we hypothesize positive improvements. The test statistic is:
\begin{equation}
    t = \frac{\bar{\Delta}}{\text{SE}(\Delta)} = \frac{\bar{\Delta}}{s_\Delta / \sqrt{n}}
\end{equation}
where $\bar{\Delta}$ is the mean improvement across $n=14$ mystery namings, $s_\Delta$ is the sample standard deviation, and $\text{SE}(\Delta)$ is the standard error.

\subsection{Results}

Table~\ref{tab:steering_improvements} shows that several steering conditions produce statistically significant improvements over baseline. Layer 20 in-naming ($p = 0.042$), layer 20 cross-naming ($p = 0.044$), and layer 40 cross-naming ($p = 0.021$) all achieve significance at $\alpha = 0.05$, with mean improvements ranging from 1.4\% to 1.8\%.

\begin{table*}[t]
\centering
\caption{Steering accuracy improvements across 14 mystery namings. SE = standard error; * $p < 0.05$, ** $p < 0.01$, *** $p < 0.001$, ns = not significant.}
\label{tab:steering_improvements}
\small
\begin{tabular}{lccccl}
\toprule
\textbf{Condition} & \textbf{Mean $\Delta$} & \textbf{SE} & \textbf{t-statistic} & \textbf{p-value} & \textbf{Sig.} \\
\midrule
Layer 20 In-naming & 1.57\% & 0.84\% & 1.878 & 0.042 & * \\
Layer 20 Cross-naming & 1.79\% & 0.97\% & 1.846 & 0.044 & * \\
Layer 20 Random Gaussian & $-0.36\%$ & 1.08\% & $-0.332$ & 0.627 & ns \\
\midrule
Layer 30 In-naming & 0.64\% & 1.29\% & 0.500 & 0.313 & ns \\
Layer 30 Cross-naming & 1.36\% & 1.20\% & 1.133 & 0.139 & ns \\
\midrule
Layer 40 In-naming & 0.57\% & 0.92\% & 0.618 & 0.274 & ns \\
Layer 40 Cross-naming & 1.43\% & 0.64\% & 2.249 & 0.021 & * \\
\midrule
Layer 50 In-naming & 0.36\% & 1.01\% & 0.352 & 0.365 & ns \\
Layer 50 Cross-naming & 0.93\% & 0.61\% & 1.531 & 0.075 & ns \\
\bottomrule
\end{tabular}
\end{table*}

Random Gaussian steering at layer 20 shows negative mean improvement ($-0.36\%$, $p = 0.627$), confirming that improvements from structured representations are not artifacts of random perturbations. Cross-naming representations show numerically higher improvements than in-naming representations at all tested layers, with the strongest and most consistent effects observed at layer 40 where cross-naming achieves 1.43\% improvement ($p = 0.021$) compared to 0.57\% for in-naming ($p = 0.274$).

\subsection{Discussion}

Our statistical analysis provides evidence that steering with refined representations improves accuracy over baseline, with the strongest effects observed at layers 20 and 40 for cross-naming steering. The significant improvements at multiple layers, combined with the lack of improvement from random Gaussian controls, support our hypothesis that representational adaptations during reasoning contain meaningful structural information that causally contributes to problem-solving performance.

The moderate effect sizes (1.4--1.8\% for significant conditions) and variability across mystery namings reflect the challenging nature of using steering with long reasoning rollouts.

\end{document}